\documentclass[journal]{IEEEtran}
\ifCLASSINFOpdf
\else
\fi
\usepackage{amsmath}
\usepackage{rotating,booktabs,multirow}
\newsavebox{\measurebox}
\usepackage{upgreek}
\usepackage{amssymb}
\usepackage{amsthm}
\usepackage{multirow}
\usepackage{mathrsfs,amsmath}
\usepackage{graphicx}
\usepackage{color}
\usepackage{xcolor}

\usepackage{mathtools}
\makeatletter

\newcommand{\Rmnum}[1]{\expandafter\@slowromancap\romannumeral #1@}
\newcommand{\tr}{^\textrm{T}}
\newcommand{\bx}{\mathbf{x}}
\newcommand{\bk}{\mathbf{k}}
\newcommand{\ba}{\mathbf{a}}

\DeclareMathOperator*{\argminA}{arg\,min}
\def\kwave~{\textbf{k}-\texttt{Wave}}

\newcommand{\todo}[1]{\textcolor{red}{#1}}

\newcommand{\dontshow}[1]{}
\newcommand{\diag}[1]{\mathrm{diag}(#1)}


\makeatletter
\newenvironment{W}[1][htb]{%
    \renewcommand{\ALG@name}{W}
   \begin{algorithm}[#1]%
  }{\end{algorithm}}
\makeatother
\makeatletter
\newenvironment{R-SALSA}[1][htb]{%
    \renewcommand{\ALG@name}{R-SALSA}
   \begin{algorithm}[#1]%
  }{\end{algorithm}}
\makeatother
\makeatletter
\newenvironment{R-FISTA}[1][htb]{%
    \renewcommand{\ALG@name}{R-FISTA}
   \begin{algorithm}[#1]%
  }{\end{algorithm}}
\makeatother
\makeatletter
\newenvironment{R-ADMM}[1][htb]{%
    \renewcommand{\ALG@name}{R-ADMM}
   \begin{algorithm}[#1]%
  }{\end{algorithm}}
\makeatother
\usepackage{algorithm,algorithmic}

\usepackage{array}

\ifCLASSOPTIONcompsoc
  \usepackage[caption=false,font=normalsize,labelfont=sf,textfont=sf]{subfig}
\else
  \usepackage[caption=false,font=footnotesize]{subfig}
\fi
\usepackage{url}

\begin{document}
%
\title{Photoacoustic Reconstruction Using Sparsity\\ in Curvelet Frame: Image versus Data Domain}
%
%
%

\author{Bolin Pan, Simon R. Arridge, Felix Lucka, Ben T. Cox, Nam Huynh, Paul C. Beard, Edward Z. Zhang and Marta M. Betcke
\thanks{Bolin Pan, Simon R. Arridge and Marta M. Betcke are with the Department of Computer Science, University College London, UK, WC1E 6BT London, UK, e-mail: m.betcke@ucl.ac.uk}
\thanks{Felix Lucka is with Department of Computational Imaging, Centrum Wiskunde \& Informatica, Netherlands, P.O. Box 94079 1090 GB Amsterdam, Netherlands.}
\thanks{Ben T. Cox and Nam Huynh and Edward Z. Zhang and Paul C. Beard are with Department of Medical Physics, University College London, UK, WC1E 6BT London, UK.}
\thanks{Nam Huynh and Edward Z. Zhang and Paul C. Beard are with Wellcome/EPSRC Centre for Interventional and Surgical Sciences, University College London, UK, WC1E 6BT London, UK.}
}

%
%

\markboth{Journal of \LaTeX\ Class Files,~Vol.~14, No.~8, August~2015}%
{Shell \MakeLowercase{\textit{et al.}}: Bare Demo of IEEEtran.cls for IEEE Journals}
%



\maketitle

\begin{abstract}
Curvelet frame is of special significance for photoacoustic tomography (PAT) due to its sparsifying and microlocalisation properties. We derive a one-to-one map between wavefront directions in image and data spaces in PAT which suggests near equivalence between the recovery of the initial pressure and PAT data from compressed/subsampled measurements when assuming sparsity in Curvelet frame. As the latter is computationally more tractable, investigation to which extent this equivalence holds conducted in this paper is of immediate practical significance. To this end we formulate and compare
\ref{DR}, a two step approach based on the recovery of the complete volume of the photoacoustic data from the subsampled data followed by the acoustic inversion, and \ref{p0R}, a one step approach where the photoacoustic image (the initial pressure, $p_0$) is directly recovered from the subsampled data. 
Effective representation of the photoacoustic data requires basis defined on the range of the photoacoustic forward operator. To this end we propose a novel \textit{wedge-restriction} of Curvelet transform which enables us to construct such basis.
Both recovery problems are formulated in a variational framework. As the Curvelet frame is heavily overdetermined, we use reweighted $\ell_1$ norm penalties to enhance the sparsity of the solution. The data reconstruction problem \ref{DR} is a standard compressed sensing recovery problem, which we solve using an ADMM-type algorithm, SALSA. Subsequently, the initial pressure is recovered using time reversal as implemented in the k-Wave Toolbox. The $p_0$ reconstruction problem, \ref{p0R}, aims to recover the photoacoustic image directly via FISTA, or ADMM when in addition including a non-negativity constraint.
We compare and discuss the relative merits of the two approaches and illustrate them on 2D simulated and 3D real data in a fair and rigorous manner.
\end{abstract}


%
\IEEEpeerreviewmaketitle

\section{Introduction}\label{sec:intro}
%
%
%
%
\IEEEPARstart{S}{parsity} is a powerful property closely related to the information content carried by a signal. Sparsity of a signal is defined with respect to a given basis\footnote{We refer to a tight frame as a normalised (overdetermined) basis}, and signals may be highly sparse in some bases while not sparse in others.
Thus the choice of the right basis is paramount to achieve effective sparse representation and hence to
the success of sparsity enhanced signal recovery also known as \emph{compressed sensing}. The theoretical underpinning of the field of compressed sensing was laid in a series of seminal papers by Cand{\`e}s, Tao \cite{NearOptimal}, Romberg \cite{Robust}, \cite{stableRecovery}, and Donoho \cite{donohoCS}. 

\subsection{Motivation}\label{sec:intro:motiv} 
This work exploits the observation that in photoacoustic tomography, similar to X-ray tomography, both the photoacoustic \emph{image} (initial pressure) and the photoacoustic \emph{data} (pressure time series recorded by the detector) exhibit sparsity in their respective domains. The photoacoustic image can be thought of as a function with singularities along $\mathcal C^2$ smooth curves for which Curvelets provide a nearly optimally sparse representation \cite{NC}. On the other hand, in the regime of geometrical optics under the action of the wave equation, Curvelets have been shown to essentially propagate along the trajectories of the underlying Hamiltonian system (the projections on the physical space of the solutions of the underlying Hamiltonian system) \cite{dataMotivation}. Assuming the initial pressure to be a single Curvelet, the wave field induced by such initial pressure at any given point in time can be approximated by simply propagating the corresponding Curvelet along the geometrical ray initiating at the centre of the Curvelet with the initial direction perpendicular to its wavefront. Consequently, if the initial pressure is sparse in the Curvelet frame the acoustic propagation effectively preserves the sparsity of the initial pressure. However, this argument does not go far enough to claim sparsity of the photoacoustic data where the acoustic wavefront group propagating in physical space ($\mathbb R^d$) is mapped to a function of time and detector coordinate (data space) via the dispersion relation given by the wave equation. In Section \ref{sec:Curvelet:WaveFrontMapping} we derive a one-to-one map between the physical space (and by Hamiltonian flow argument also the image space) and the data space therewith essentially inducing the same sparsity of the initial pressure onto the photoacoustic data. Motivated by these results, we hypothesise near equivalence
of the sparse representation of both the initial pressure (photoacoustic image) and the photoacoustic data in Curvelet frame.

%
%
\subsection{Contribution}\label{sec:intro:cont}
In this paper we focus on the problem of image reconstruction from subsampled photoacoustic data measured using a Fabry-P\'erot planar detector scanner. An example of such a system is the single point multiwavelength backward-mode scanner \cite{PATsys, SubsampledFP17}. The proposed methodology also directly applies to compressed measurements, provided the sensing patterns form a (scrambled) unitary transform e.g.~scrambled Hadamard transform. A PAT system capable of compressed measurements acquisition was introduced in \cite{PatternPAT14,RTUSField15,SPOC}.   

As outlined in the motivation, we argue the near optimal sparsity of the Curvelet representation of both the entire volume of the photoacoustic data and the photoacoustic image (initial pressure) and stipulate a near equivalence of both recovery problems based on a one-to-one wavefront direction map which we derive.
To capitalise on the sparsity of data, we deploy a \emph{two step approach} which first solves a basis pursuit problem to recover the full photoacoustic data volume from subsampled/compressed measurements, followed by a standard acoustic inversion to obtain the initial pressure (\ref{DR}).
We demonstrate that efficient and robust representation of the photoacoustic data volume requires basis which is defined on the range of the forward operator and propose a novel \textit{wedge-restriction} of the standard Curvelet transform which allows us to construct such basis. On the other hand, motivated by the sparsity of the photoacoustic image in the Curvelet frame, we consider a \emph{one step approach}, which recovers the photoacoustic image $p_0$ directly from the subsampled measurements (\ref{p0R}).\footnote{A flow chart diagram can be found in supplementary material} In contrast to prior work \cite{Acoustic}, our new data reconstruction method builds upon the one-to-one wavefront direction map and captures the relationship between the time-steps. Therefore, it has the potential to perform on a par with the \ref{p0R} approach \textit{without the need to iterate with the expensive forward operator}. To verify this hypothesis
we rigorously compare these two strategies on an appropriately constructed simulated 2D example and real 3D PAT data. In both cases the sparse recovery is formulated in a variational framework, utilising an iteratively reweighted $\ell_1$ norm to more aggressively pursue sparsity in an overdetermined frame, as appropriate for the Curvelets basis used herein. We discuss appropriate choices of algorithms for the solution of the two resulting optimisation problems. Finally, while we focus the presentation on photoacoustic tomography, the methodology applies to other imaging modalities where the forward operator admits microlocal characterization, most prominently X-ray computed tomography.

%
%
\subsection{Related work}\label{sec:intro:prior}
A number of one and two step approaches have been previously proposed in the context of photoacoustic tomography. To the best of our knowledge \cite{ACSPAT} presents the first one step approach for a circular geometry with focused transducers for 2D imaging using sparse representation in Wavelets, Curvelets (with Wavelets at the finest scale), and Fourier domains, and first derivative (total variation). 
The 3D reconstruction problem for PAT with an array of ultrasonic transducers using sparsity in Wavelet frame was presented in \cite{CSPATinvivo}. An iterative reconstruction approach based on an algebraic adjoint with sparse sampling for 2D and 3D photoacoustic images was first introduced in \cite{FullWave}. Two analytical adjoints have been subsequently proposed, one in a BEM-FEM setting \cite{FBEMadjoint} and the other in ${\bf k}$-space setting \cite{PATadjoint}, and exploited/refined in a series of papers \cite{accelerate,javaherian2016multi,haltmeier2017analysis}. The one step approach in the present work builds upon \cite{PATadjoint}, \cite{accelerate} and extends it to sparsity in the Curvelet frame in 2D and 3D. 
\par

The two step approach suggested in \cite{Acoustic} recovers the photoacoustic data in every time step independently from pattern measurements under assumption of sparsity of the photoacoustic pressure on the detector at a fixed time in a 2D Curvelet frame. It has the advantage that each of the  
time step recovery problems can be solved independently, allowing trivial parallelization. The disadvantage is that the correspondence between the time steps is lost: \emph{i.e.}~the recovery does not take into account that the photoacoustic data corresponds to the solution to the same wave equation evaluated at the detector at different time steps. In another approach proposed in \cite{CSSPAT} the authors explore the temporal sparsity of the entire photoacoustic data time series using a custom made transform in time. Then the initial pressure is reconstructed via the universal back-projection (UBP) method \cite{xu2005universal}.

In the two step approach proposed here, we use a variational formulation akin to the one in \cite{Acoustic} while we represent the entire volume of data, \emph{i.e.}~sensor location and time, akin to \cite{CSSPAT}, but in the Curvelet basis. This approach combines the best of both worlds: representing the entire data volume naturally preserves the connection between the time steps, while, as indicated above and discussed in Section \ref{sec:Curvelet:RestrictedCurvelet}, essentially optimal data sparsity is achieved using such 2D/3D Curvelet representation.

%
%
\subsection{Outline}\label{sec:intro:out}

The remainder of this paper is structured as follows. In Section \ref{sec:PAT} we provide the background and formulation of the forward, adjoint and inverse problems in PAT. Section \ref{sec:Curvelet} discusses the details of sparse Curvelet representation of PAT image and PAT data. First, we derive a one-to-one map between wavefronts in ambient space (and hence image space) and data space which motivates use of Curvelet transform also for the representation of the PAT data volume. Then we briefly recall the Curvelet transform and propose a concept of \textit{wedge restriction} which allows us to formulate a tight frame of \textit{wedge restricted} Curvelets on the range of the PAT forward operator for PAT data representation. Finally, we consider details specific to PAT data such as symmetries, time oversampling and their consequences. In Section \ref{sec:Optimization} we formulate the reconstruction problem from subsampled measurements via \ref{DR} and \ref{p0R} approaches in a variational framework, and propose tailored algorithms for their solution. In Section \ref{sec:PATRec} the relative performance of both approaches is rigorously evaluated on a specially constructed 2D example and on experimental 3D real data of a human palm vasculature. We also compare our one step approach using Curvelets with one step approach using Wavelet bases and against state of the art total variation regularisation. We evaluate robustness of both one and two step approaches to noise and more aggressive subsampling. We summarize the results of our work and point out future research directions in Section \ref{sec:Conclusion}.

%
%
\section{Photoacoustic Tomography}\label{sec:PAT}

Photoacoustic tomography (PAT) has been steadily gaining in importance for biomedical and preclinical applications. The fundamental advantage of the technique is that optical contrast is encoded on to acoustic waves, which are scattered much less than photons. It therefore avoids the depth and spatial resolution limitations of purely optical imaging techniques that arise due to strong light scattering in tissue: with PAT, depths of a few centimetres with scalable spatial resolution ranging from tens to hundreds of microns (depending on depth) are achievable. However, obtaining high resolution 3D PAT images with a Nyquist limited spatial resolution of the order of 100 $\upmu$m requires spatial sampling intervals on a scale of tens of $\upmu$m over cm scale apertures. This in turn necessitates recording the photoacoustic waves at several thousand different spatial points which results in excessively long acquisition times using sequential scanning schemes, such as the Fabry--P{\'{e}}rot based PA scanner, or mechanically scanned piezoelectric receivers. In principle, this can be overcome by using an array of detectors. However, a fully sampled array comprising several thousand detector elements would be prohibitively expensive. Hence, new methods that can reconstruct an accurate image using only a subset of this data would be highly beneficial in terms of acquisition speed or cost.

A PAT system utilises a near infrared laser pulse which is delivered into biological tissue. Upon delivery some of the deposited energy will be absorbed and converted into heat which is further converted to pressure. This pressure is then released in the form of acoustic pressure waves propagating outwards. 
When this wave impinges on the ultrasound detector $\mathcal{S} \subset \mathbb{R}^d$, it is recorded at a set of locations $\bx_{\mathcal S} \in \mathcal{S}$ over time constituting the photoacoustic data $g(t,\bx_{\mathcal S})$ which is subsequently used for recovery of the initial pressure $p_0$.

With several assumptions on the tissues properties \cite{wang2011photoacoustic}, the photoacoustic forward problem can be modelled as an initial value problem for the free space wave equation
\begin{equation}
\label{eq:PATforwardequation}\tag{{\bf A}}
\begin{aligned}
\nonumber \underbrace{\left( \frac{1}{c^2(\bx)}\frac{\partial^2}{\partial t^2}  - \nabla^2 \right)}_{:=\square^2 } p(t,\bx) &= 0, \quad (t,\bx)\in (0,T)\times\mathbb{R}^d, \\
p(0,\bx) & = p_0(\bx), \;
p_t(0,\bx)  = 0,
\end{aligned}
\end{equation}
where $p(t,\bx)$ is time dependent acoustic pressure in $\mathcal{C}^\infty((0,T)\times\mathbb{R}^d)$ and $c(\bx) \in \mathcal{C}^\infty(\Omega)$ is the speed of sound in the tissue.\par
The photoacoustic inverse problem recovers initial pressure $p_0(\bx) \in \mathcal{C}_0^\infty(\Omega)$ in the region of interest $\Omega$ on which $p_0(\bx)$ is compactly supported, from time dependent measurements 
\begin{equation*}
\setlength\abovedisplayskip{5pt}
\setlength\belowdisplayskip{5pt}    
g(t,\bx_{\mathcal{S}}) = \omega(t) p(t,\bx), \quad (t,\bx_{\mathcal{S}})\in (0,T)\times \mathcal S
\end{equation*} 
on a surface $\mathcal S$, e.g.~the boundary of $\Omega$, where $\omega\in \mathcal{C}_0^{\infty}(0,T)$ is a temporal smooth cut-off function. It amounts to a solution of the following initial value problem for the wave equation with constraints on the surface \cite{PATbackward}
\begin{equation}\label{eq:TRequation}\tag{\bf TR}
\begin{aligned}
\square^2 p(t,\bx) &=0, & \text{}  (t,\bx)\in (0,T)\times\Omega, \\
p(0,\bx) & = 0, \; p_t(0,\bx) = 0, & \text{}  \\
p(t,\bx) & = g(T-t,\bx_{\mathcal{S}}), & \text{}(t,\bx_{\mathcal{S}})\in (0,T)\times\mathcal S.
\end{aligned}
\end{equation}
also referred to as a \emph{time reversal}.
For non-trapping smooth $c(\bx)$, in 3D, the solution of \eqref{eq:TRequation} is the solution of the PAT inverse problems if $T$ is chosen large enough so that $g(t,\bx) = 0$ for all $\bx\in\Omega, t\geq T$ and the wave has left the domain $\Omega$. Assuming that the measurement surface $\mathcal S$ surrounds the region of interest $\Omega$ containing the support of initial pressure $p_0$, the wave equation \eqref{eq:TRequation} has a unique solution. When no complete data is available (meaning that some singularities have not been observed at all as opposed to  the case where partial energy has been recorded, see e.g.~\cite{StefanovUhlman}), $p_0$ is usually recovered in a variational framework, including some additional penalty functional corresponding to prior knowledge about the solution. Iterative methods for solution of such problems require application of the adjoint operator which amounts to the solution of the following initial value problem with time varying mass source \cite{PATadjoint}
\begin{equation}\label{eq:adjoint}\tag{$\textbf{A}^\star$}
\begin{aligned}
\square^2 q(t,\bx) &= \left\{
\begin{array}{cc} 
\frac{\partial}{\partial t} g(T-t,\bx_{\mathcal{S}}), &(t,\bx_{\mathcal{S}})\in  (0,T)\times\mathcal S\\ 
0 & \text{everywhere else} 
\end{array}
\right.\\
q(0,\bx) & = 0, \; q_t(0,\bx) = 0 & \text{ }
\end{aligned}
\end{equation}
evaluated at $T$, $q(T,\bx)$.
All the equations \eqref{eq:PATforwardequation}, \eqref{eq:TRequation} and \eqref{eq:adjoint} can be solved using the pseudospectral method implemented in the k-Wave toolbox for heterogeneous media in two and three dimensions \cite{kWave}; see \cite{PATadjoint} for implementation of the adjoint \eqref{eq:adjoint} 
in this setting.\par

%
\section{Multiscale Representation of Photoacoustic Data and Initial Pressure}\label{sec:Curvelet}
The presentation in this section assumes a cuboid domain $\Omega$ with a flat panel sensor $\mathcal S$ at one of its sides and constant speed of sound $c(\bx) = c$.

%
\subsection{Wavefront direction mapping}\label{sec:Curvelet:WaveFrontMapping}

The solution of wave equation \eqref{eq:PATforwardequation} for a constant speed of sound $c(\mathbf{x}) = c$, assuming half space symmetry as per planar sensor geometry,
can be explicitly written in Fourier domain as \cite{cox2005fast}
\begin{equation}\label{eq:p0cos}
\hat p(t,\bk) = \cos(c|\mathbf{k}|t)\hat  p_0(\bk),
\end{equation}
where $\hat p$ denotes the Fourier transform of $p$ (in all or some variables) and $\mathbf{k}$ is the Fourier domain wave vector.

\begin{figure}[ht]
\vspace{-0.4cm} 
\setlength{\abovecaptionskip}{-0.1cm} 
  \centering
  \includegraphics[width=0.5\linewidth]{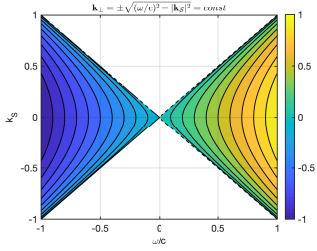}
  \caption[bowtie]{Contour plot of $\bk_\perp=\mbox{sign}(\omega)\sqrt{(\omega/c)^2-|\mathbf{k}_{\mathcal{S}}|^2} = const$ over the bow-tie shaped frequency domain range of the photoacoustic forward operator, 
  $\mathcal R_A = \left\{ (\omega/c, \bk_{\mathcal S}) \in \mathbb R^d: \omega/c>|\mathbf{k}_{\mathcal{S}}|\right\} \cup {\bf 0}$.}
  \label{image:bowtie}
\end{figure}

Using \eqref{eq:p0cos}, the following mapping can be obtained between the initial pressure $p_0$ and the PAT data $p(ct, \bx_{\mathcal S})$ in Fourier domain (note we use $c$ normalised time/frequencies, the  non-normalised formulation contains additional factor $1/c$ on the right hand side) \cite{TBPO,EFRFTT,Acoustic}
\begin{equation}\label{eq:frequencyEquation}
\hat p(\omega/c, \bk_{\mathcal{S}}) = \frac{\omega/c}{\sqrt{(\omega/c)^2-{|\mathbf{k}_{\mathcal{S}}}|^2}} \hat p_0\left(\sqrt{ (\omega/c)^2- |\mathbf{k}_{\mathcal{S}}}|^2, \bk_{\mathcal{S}}\right). 
\end{equation}
Equation \eqref{eq:frequencyEquation} describes change from Cartesian to hyperbolic coordinates including scaling by the Jacobian determinant. 
For this mapping to be real and bounded, the expression under the square root has to be positive bounded away from 0 for all non-zero wave vectors, \emph{i.e.}~$\mathcal R_A = \left\{ (\omega/c, \bk_{\mathcal S})  \in \mathbb R^d: \omega/c>|\mathbf{k}_{\mathcal{S}}|\right\} \cup {\bf 0}$,  resulting in a bow-tie shaped range of the photoacoustic operator in the sound-speed normalised Fourier space as illustrated by the contour plot of $\bk_\perp=\mbox{sign}(\omega)\sqrt{(\omega/c)^2-|\mathbf{k}_{\mathcal{S}}|^2}$, $\bk_\perp = const$, in Figure \ref{image:bowtie}.

The change of coordinates described by equation \eqref{eq:frequencyEquation} (which is a consequence of the dispersion relation in the wave equation $(\omega/c)^2 = |\bk|^2 = |\bk_{\mathcal S}|^2 + |\bk_{\perp}|^2$) defines a one-to-one map between the frequency vectors $\bk \in \mathbb R^d$ and $( \omega/c, \bk_{\mathcal S}) \in \mathcal R_A$ 
\begin{align}\label{eq:map:k}
\bk = (\bk_\perp, \bk_{\mathcal S} ) &\leftarrow \left( \sqrt {(\omega/c)^2 - |\bk_{\mathcal S}|^2 }, \bk_{\mathcal S} \right), \\
\nonumber (\omega/c, \bk_{\mathcal S}) &\leftarrow \left(|\bk| = \sqrt { |\bk_\perp|^2 + |\bk_{\mathcal S}|^2 }, \bk_{\mathcal S} \right).
\end{align}
This mapping is illustrated below. Figure \ref{image:PATforwardmodel} shows the wavefronts and the corresponding wavefront vectors in the ambient space $\bar{\theta} = (-\cos\theta, \sin\theta)$ on the left and in the data space $\bar{\beta} = (\cos\beta, -\sin\beta)$ on the right. Note that these wavefront vectors have the same direction as their frequency domain counterparts $\bk_{\bar\theta}, \bk_{\bar\beta}$ which are depicted in Figure \ref{image:PATforwardmodel:k}.

\par
\begin{figure}[ht]
\vspace{-0.4cm} 
\setlength{\abovecaptionskip}{-0.1cm} 
\setlength{\belowcaptionskip}{-1cm}   
  \centering
  \includegraphics[width=0.8\linewidth,height=0.48\linewidth]{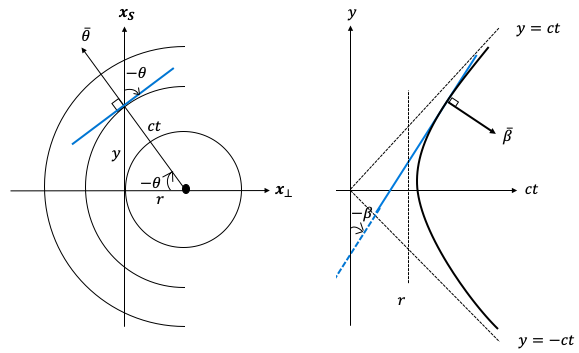}
  \caption{Wavefront direction mapping between the ambient space $\bar\theta$ (left), and the data space, $\bar\beta$ (right) on an example of spherical wave in 2D (which can be seen as a cross-section of a 3D problem with rotational symmetry around $ct$ axis).
  $-\theta$ and $-\beta$ are the angles that the wavefronts make with the sensor plane. Their signs follow the convention of right handed coordinate system $(\bx_\perp, \bx_{\mathcal S})$.
  The respective wavefronts (their tangent planes) are depicted in blue.}
  \label{image:PATforwardmodel}
\end{figure}
\begin{figure}[ht]
\vspace{-0.7cm} 
\setlength{\abovecaptionskip}{-0.1cm} 
  \centering
  \includegraphics[width=\linewidth,height=0.4\linewidth]{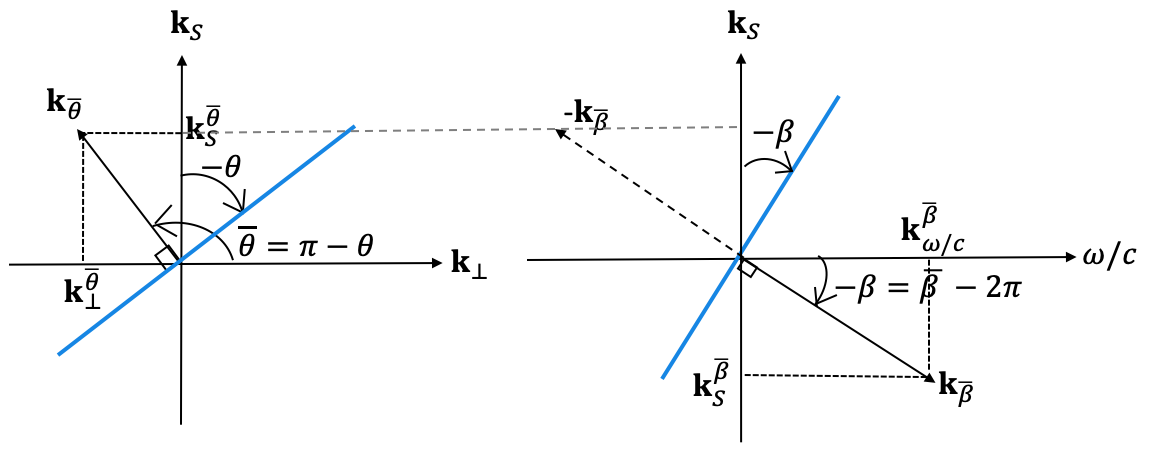}
  \caption{Frequency domain visualisation of the example in Figure \ref{image:PATforwardmodel}. Fourier domain wavefront vector mapping between the ambient space, $\bk_{\bar\theta}$ (left), and the data space, $\bk_{\bar\beta}$ (right).}
  \label{image:PATforwardmodel:k}
\end{figure}

Formula \eqref{eq:frequencyEquation} has an implicit assumption of even symmetry of $p_0(\bx_\perp, \bx_\mathcal S ), \, p( ct, \bx_\mathcal S)$ in $\mathbb R^d$ w.r.t.~$\bx_\perp = 0, \, ct = 0$, respectively, which is inherited by the map \eqref{eq:map:k}
\begin{align}\label{eq:map:k:mir}
\hat p_0 (-\bk_\perp, \bk_{\mathcal S})  &= \hat p_0  ( \bk_\perp, \bk_{\mathcal S}), \\
\nonumber \hat p  (-\omega/c, \bk_{\mathcal S}) &= \hat p  (\omega/c, \bk_{\mathcal S} ).
\end{align}
We note, that the even symmetry mirrors the spherical wave source (and consequently the data) in Figures \ref{image:PATforwardmodel}, \ref{image:PATforwardmodel:k} w.r.t.~the sensor plane. As a result we actually have two wavefronts synchronously impinging at the same point of the detector with angles $-\theta$ (original $p_0$) and $\theta$ (mirrored $p_0$). The first is mapped to wavefront with angle $-\beta$ (original data $p( \omega/c,\bx_{\mathcal S})$) and $\beta$ (mirrored data $p(-\omega/c,\bx_{\mathcal S} )$).

In contrast, the symmetric Curvelet transform introduced in Section
\ref{sec:Curvelet:CurveletTransform} below assumes Hermitian symmetry which holds for Fourier transforms of real functions. Thus computing symmetric Curvelet transform of un-mirrored real functions\footnote{To keep the notation light, we do not explicitly distinguish between mirrored and un-mirrored quantities as it should be clear from the context.}: initial pressure $p_0( \bx_{\perp}, \bx_{\mathcal S})$ (supported on $\bx_{\perp} \geq 0$ and 0 otherwise) and data $p(\omega/c,\bx_{\mathcal S})$ (supported on $\omega/c \geq 0$ and 0 otherwise), we instead continue the map \eqref{eq:map:k} with Hermitian symmetry into the whole of $\mathbb R^d$
\begin{align}\label{eq:map:k:mirconConj}
\hat p_0 (-\bk_\perp,\bk_{\mathcal S})  &= \mbox{Conj}(\hat p_0  (\bk_\perp, -\bk_{\mathcal S})) \\
\nonumber \hat p (-\omega/c, \bk_{\mathcal S}) &= \mbox{Conj}(\hat p (\omega/c,-\bk_{\mathcal S})).
\end{align}
Here $\mbox{Conj}(\cdot)$ denotes complex conjugate. 
The same Hermitian symmetry underpins the wavefront vectors which orientation is arbitrary i.e.~$\bk_{\bar\theta}$ vs.~$-\bk_{\bar\theta}$ or $\bk_{\bar\beta}$ vs.~$-\bk_{\bar\beta}$ (the latter is highlighted in Figure \ref{image:PATforwardmodel:k}(right)). 
We note, that neither the mirroring nor the Hermitian symmetry affect the wavefront angle. While mirroring duplicates the wavefronts in a sense explained above, this is not the case for Hermitian symmetry as the wavefront vector is defined up to the Hermitian symmetry. In conclusion, both formulations coincide on $\omega/c \geq 0, \bk_\perp \geq 0$ and consequently wavefront angle mapping derived from mirrored formulation holds valid for the un-mirrored formulation on $\omega/c \geq 0, \bk_\perp \geq 0$ but with Hermitian instead of even symmetry.

With the notation in Figures \ref{image:PATforwardmodel} and \ref{image:PATforwardmodel:k} we immediately write the map between the wavefront vector angles\footnote{For simplicity we use $\bar\theta, \bar\beta$ to represent both the spatial domain vectors and their angles depending on the context.} (which are the same in space and frequency domains)
\begin{align}\label{eq:map:kangles}
\setlength\abovedisplayskip{3pt}
\setlength\belowdisplayskip{3pt}  
\tan \bar\beta &= \frac{\bk_{\mathcal S}^{\bar\beta}}{\bk_{\omega/c}^{\bar\beta}}= \frac{-\bk_{\mathcal S}^{\bar\theta}}{|\bk^{\bar\theta}| =\sqrt{(|\bk_\perp^{\bar\theta}|^2 + |\bk_{\mathcal S}^{\bar\theta}|^2 }} = \sin(-\bar\theta).
\end{align}  
Expressing \eqref{eq:map:kangles} in terms of the detector impingement angles using their relation with wavefront vector angles $\bar\theta = \pi -\theta$ and $\bar\beta = 2\pi -\beta$ and basic trigonometric identities we obtain the corresponding map between $\theta$ and $\beta$ (note that actual impingement angles are negatives) 
\begin{equation}\label{eq:WaveFrontMapping}
\setlength\abovedisplayskip{3pt}
\setlength\belowdisplayskip{3pt}  
\beta = \arctan\left(\sin\theta \right).
\end{equation}
While the derivation was done in 2D for clarity it is easily lifted to 3D. In 3D we have two angles describing the wavefront in ambient space $\theta = (\theta^1, \theta^2)\tr$ (in planes spanned by $\bk_\perp$ and each detector coordinate axis $\hat\bk_{\mathcal S}^{i}$) and analogously in data space $\beta = (\beta^1, \beta^2)\tr$, with the map \eqref{eq:WaveFrontMapping} and all statements holding component wise.

We would like to highlight that this map only depends on the angle $\theta$ (not the distance $r$ nor translation along the detector). The wavefront direction mapping is universal across the entire data volume regardless of the position of the source w.r.t.~the detector. For the un-mirrored initial pressure, as $\theta$ sweeps the angle range $(-\pi/2, \pi/2)$, $\beta$ sweeps $\left(-\frac{\pi}{4}, \frac{\pi}{4}\right)$. For mirrored initial pressure, these angles can be simply negated which corresponds to mirroring them around the sensor plane. This is consistent with the hyperbolic envelope, which restricts the range of the PAT forward operator in frequency domain to a bow-tie region $\mathcal R_A$. We will make use of this observation in Section \ref{sec:Curvelet:RestrictedCurvelet} for Curvelet representation of the PAT data.

%
%
\subsection{Curvelets}\label{sec:Curvelet:CurveletTransform}
\begin{figure}[htbp!] 
\vspace{-0.4cm} 
\setlength{\abovecaptionskip}{-0.2cm} 
\setlength{\belowcaptionskip}{-0.1cm}   
\centering    
\includegraphics[width=0.65\linewidth]{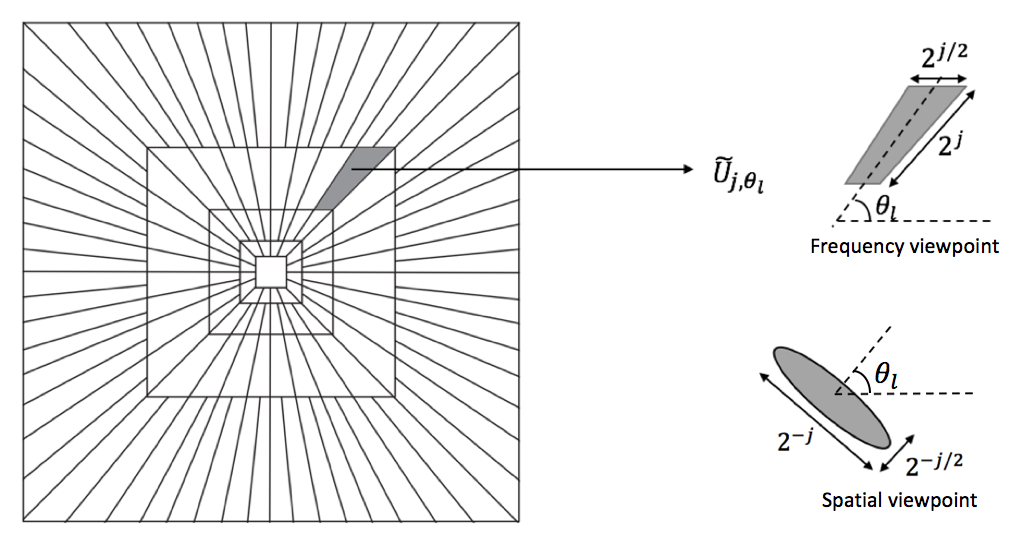}
\caption[2DTilling]{An example of 2D frequency domain Curvelet tiling using 5 scales. The highlighted wedge, magnified view in top right, corresponds to the
frequency window near which the Curvelet $\tilde{U}_{j,\theta_l}$
at the scale $j$ with orientation $\theta_l$ is supported. The orientation of the envelope of the corresponding Curvelet function in the spatial domain is shown in bottom right corner.}
\label{image:2DTilling}
\end{figure}

\dontshow{
Sparse representation in the Fourier domain fails to provide correspondence between frequency content and spatial coordinates. The Wavelet transform attempts to overcome this problem by dividing the signal into different frequency bands and generating sets of coefficients where each set contains the spatial content relevant to a corresponding frequency band. However, Wavelet transform is not well adapted to represent directional singularities such as edges or wavefronts. On the other hand, the Curvelet transform \cite{NC}, \cite{FDCT} belongs to a family of transforms which were designed with this purpose in mind \cite{candes1998ridgelets}, \cite{candes1999ridgelets}, \cite{antoine1996one}, \cite{antoine1996two}, \cite{kutyniok2012shearlets}, \cite{kutyniok2016shearlab}, \cite{do2005contourlet},\dontshow{\todo{include citations to Ridgelets, directional wavelets, Shearlets, Countourlets}} and is the representation of choice for the present work.
\par
}
Curvelet transform is a multiscale pyramid with different directions and positions. Figure \ref{image:2DTilling} 
exemplifies a typical 2D Curvelet tiling of the frequency domain underpinning the transform introduced in \cite{FDCT}. For each scale $j \geq j_0$ and angle $\theta_l$, $\theta_l \in [-\pi,\pi)$, in spatial domain 
the Curvelet envelope $\tilde{U}_{j,\theta_l}(\bx)$ is aligned along a ridge of length $2^{-j/2}$ and width $2^{-j}$. In the frequency domain, a Curvelet is supported near a trapezoidal wedge with the orientation $\theta_l$. The Curvelet wedges become finer with increasing scales which makes them capable of resolving singularities along curves. 

Here we introduce Curvelet transform straight away in $\mathbb R^d$ following the continuous presentation of the Curvelet transform in \cite{FDCT}.
Utilising Plancherel’s theorem, Curvelet transform is computed in frequency domain. For each scale $j$, orientation $\theta_l =  2\pi/L \cdot 2^{-\lfloor j/2 \rfloor} \cdot l , l \in \mathbb Z^{d-1}$ 
such that $\theta_l \in [-\pi,\pi)^{d-1}$ ($L \cdot 2^{\lfloor j_0/2 \rfloor}$ is the number of angles at the second coarsest scale $j=j_0+1$, $j_0$ even), and translation ${\bf a} \in \mathbb Z^d$, the Curvelet coefficients of $u: \mathbb{R}^d\rightarrow \mathbb R$, are computed as a projection of its Fourier transform $\hat{u}(\mathbf{k})$ on the corresponding basis functions $\tilde{U}_{j,\theta_l}(\bk) \exp(-i\bk \cdot \bx_{\ba}^{(j,l)})$ with $\tilde{U}_{j,\theta_l}({\bf k})$ a trapezoidal frequency window with scale $j$ and orientation $\theta_l$, and spatial domain centre at $\bx_{\ba}^{(j,l)}$ corresponding to the sequence of translation parameters ${\bf a} = (a_1,\dots, a_d)\in \mathbb{Z}^d\,:\, {\bf a} = (a_1 \cdot 2^{-j}, a_2 \cdot 2^{-j/2}, \dots, a_d \cdot 2^{-j/2})$
\begin{equation}\label{eq:NonwrapCurveletformula}\tag{C}
\setlength\abovedisplayskip{3pt}
\setlength\belowdisplayskip{3pt}  
C_{j,l}({\bf a}) = \int_{\mathbb R^d} \hat{u}({\bf k})\tilde{U}_{j,\theta_l}({\bf k}) e^{i {\bx_{\ba}^{(j,l)}} \cdot \bk} d{\bf k}.
\end{equation}
We would like to highlight the effect of the parabolic scaling on the translation grid which is reflected in the first dimension being scaled differently $a_1 \cdot 2^{-j}$ than the remaining dimensions $a_i \cdot 2^{-j/2}, i=2,\dots, d$ (the spatial translation grid is denser along the dimension 1) which is of significance for the wedge restricted Curvelet transform proposed in the next section. As we are only interested in representing real functions $u$, we use the symmetrised version of the Curvelet transform which forces the symmetry (w.r.t.~origin) of the corresponding frequency domain wedges $\theta_l$ and $\theta_l + \pi$. The coarse scale Curvelets are isotropic Wavelets corresponding to translations of an isotropic low pass window $\tilde U_0(\bk)$. The window functions at all scales and orientations form a smooth partition of unity on the relevant part of $\mathbb R^d$.

There are different ways to numerically efficiently evaluate the integral. In this paper, we use the digital coronisation with shear rather than rotation and its implementation via wrapping introduced in \cite{FDCT}. The resulting discrete transform $\Psi\, : \mathbb R^n \rightarrow \mathbb C^N$ (or $\mathbb R^N$ when symmetric) has a finite number of discrete angles $\theta_\ell$ per scale (refined every second scale according to parabolic scaling) and a fixed (per quadrant and scale) finite grid of translations $\bx_{\bf a}^{(j)} = (a_1 \cdot 2^{-j}, a_2 \cdot 2^{-j/2}, \dots, a_d \cdot 2^{-j/2})$ as opposed to the continuous definition where the grids are rotated to align with the Curvelet angle.
This fast digital Curvelet transform is a numerical isometry. For more details we refer to \cite{FDCT} and the 2D and 3D implementations in the Curvelab package\footnote{http://www.curvelet.org/software.html}.\par
\begin{figure}[htbp!]
\vspace{-0.4cm} 
\setlength{\abovecaptionskip}{0.1cm} 
\setlength{\belowcaptionskip}{-0.5cm}   
\centering
  \subfloat[$C_{4,7}((54,37))$]{
  \includegraphics[height=0.25\linewidth]{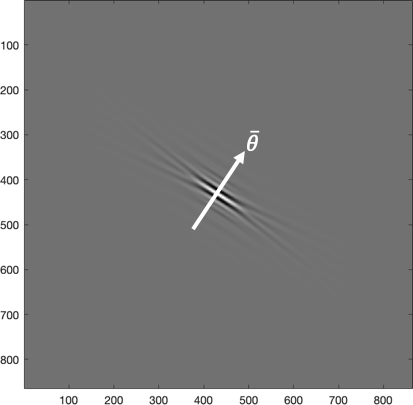}}
  \hspace{1.2pt}
  \subfloat[$\tilde{C}_{4,13}((26, 75))$]{
  \includegraphics[height=0.25\linewidth]{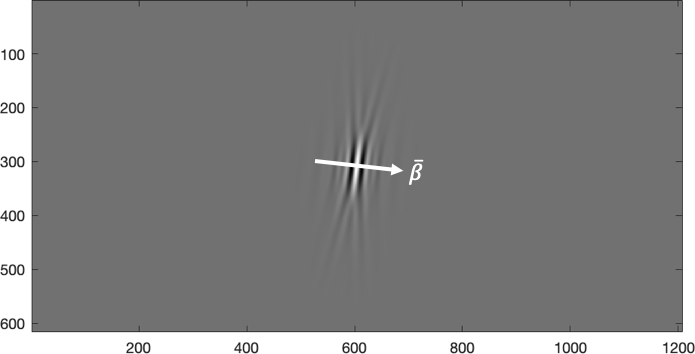}}
  \caption{An example of a single Curvelet (a) in image domain, $C_{4,7}((54,37))$, (b) and a corresponding one in the data domain, $\tilde{C}_{4,13}((26, 75))$.}
  \label{image:InvertCurvelet}
\end{figure}
\dontshow{However, the window defined $\tilde{U}_{j,\theta_l}(\omega)$ in the frequency plane does not fit in a rectangle of size $2^j \times 2^{j/2}$ directly. A solution to remedy this was presented in \cite{FDCT} which is called the wrapping approach. The wrapping $W_j$ at scale $j$ is done by choosing a parallelogram with size $2^j \times 2^{j/2}$ that contains the frequency support of the Curvelet, then periodically wrapping the product $\hat{g}(\omega)\tilde{U}_{j,\theta_l}(\omega)$ around the origin and collecting the rectangular coefficient area in the centre. One can rewrite \eqref{eq:NonwrapCurveletformula} to compute the Curvelet coefficients via wrapping 
\begin{equation}
f(j,l,k) = \int W_j(\hat{g}(\omega)\tilde{U}_{j,\theta_l}(\omega)) e^{i\langle b,\omega\rangle}d\omega.
\end{equation}
Wrapping based Curvelet transform has been shown to be a numerical isometry \cite{FDCT}. 
The 2D Curvelet transform via wrapping has been extended to 3D while preserving the same properties \cite{3DFDCT}.\par
}

%
%
\subsection{Wedge restricted Curvelets}\label{sec:Curvelet:wrc}
We now introduce a modification of the Curvelet transform to a subset of orientations in an abstract setting first, before we tailor it to represent the PAT data in the next section. We recall that the Curvelet orientation is parametrised by a vector of angles $\theta \in \mathbb [-\pi, \pi)^{d-1}$. 
Let $\bk_\theta$ denote the Curvelet wavefront vector and $\hat\bk^i$ the vector along the $i$th coordinate direction.   
Then the components of $\theta$ are angles $\theta^i, i=1,\dots,d-1$ between $\hat\bk^1$ and $\bk_\theta$ measured in planes spanned by $\hat\bk^1$ and each of the remaining coordinate vectors $\hat\bk^{i+1}, i=1,\dots,d-1$, respectively. $\theta={\bf 0}$ corresponds to Curvelets with orientation $\bk_\theta = \hat\bk^1$.
In 2D $\theta \in [-\pi, \pi)$ is just a single angle $\angle (\hat\bk^1, \bk_\theta)$ and in 3D we have two angles.

With this notation the \textit{wedge restricted Curvelet transform} amounts to computing the Curvelet coefficients i.e.~evaluating \eqref{eq:NonwrapCurveletformula} only for a wedge-shaped subset of angles $\theta_l \in \mathbb [\theta_b, \theta^t] \subset [-\pi, \pi)^{d-1}$. We furthermore define a symmetrised \textit{wedge restricted Curvelet transform} by restricting the Curvelet orientations to the symmetric double wedge (bow-tie) $\theta_l \in \mathbb [-\theta^w, \theta^w] \cup [-\theta^w+\pi, \theta^w+\pi]$ with $\theta^w \in (0, \pi/2)^{d-1}$. An example of a symmetric wedge restricted transform is illustrated in Figure \ref{image:2Dgrey} for 2D (a) and 3D (c), where the grey region corresponds to the Curvelet orientations inside the symmetric double wedge and the white to orientations outside it which are excluded from the computation.

Formally, we can write the wedge restricted Curvelet transform $\tilde{\Psi}: \mathbb R^{n} \rightarrow \mathbb R^{\tilde N}, \; \tilde N < N$ as a composition of the standard Curvelet transform $\Psi : \mathbb R^{n} \rightarrow \mathbb R^{N}$ and a projection operator $P_W : \mathbb R^{N} \rightarrow \mathbb R^{\tilde N}$, $\tilde{\Psi} = P_W \Psi$, where $P_W$ projects on the directions inside the 
bow-tie shaped set $W = [-\theta^w, \theta^w] \cup [-\theta^w+\pi, \theta^w+\pi]$ with $\theta^w \in (0, \pi/2)^{d-1}$. 
As the coarse scale, $j=j_0$, Curvelets are isotropic Wavelets, they are non-directional and thus are mapped to themselves. 
Here, we focused on the symmetric transform but definitions for complex Curvelets follow analogously.

Bearing in mind the definition of the (digital) Curvelet transform where the angles $\theta_{j,l}$ are quantised to a number of discrete orientations (which essentially correspond to the centres of the wedges), the projector $P_W$
maps all Curvelets $C_{j,l}$ with orientations $\theta_{j,l}\in W$ to themselves and all $C_{j,l}$ with orientations $\theta_{j,l} \not\in W$ to $\mathbf{0} \in \mathbb R^{\tilde N}$
\begin{align}
\setlength\abovedisplayskip{3pt}
\setlength\belowdisplayskip{3pt}  
\label{eq:projR}
&P_W: \mathbb{R}^N \rightarrow \mathbb{R}^{\tilde N}\\
\nonumber &P_W(C_{j,l}(\mathbf{a})) =
\begin{cases}
C_{j,l}(\mathbf{a}),  & \theta_{j,l}\in W, \, j > j_0\\
C_{j,l}(\mathbf{a}),  & j = j_0\\
\mathbf{0},  & \text{otherwise}.
\end{cases}
\end{align}
We note that the underlying full Curvelet transform is an isometry i.e.~its left inverse $\Psi^\dagger = \Psi^{\rm T}$. The same holds for the wedge restricted transform when defined on the restriction to the range of $P_W$ i.e.~ 
${\tilde \Psi}\, : \left.\mathbb R^n\right|_{\textrm{range}(P_W)} \rightarrow \mathbb R^{\tilde N}$
(assuming the resulting $\tilde\Psi$ is over-determined) as, by definition, on its range $P_W$ is an identity operator. 
The range of $P_W$ corresponds to the set $W \cup \tilde U_{0}$ in the Fourier domain, union of the wedge $W$ and $\tilde U_{0}$ the low frequency window at coarse scale $j_0$. Intuitively, the restriction to range of $P_W$, $\left.\mathbb R^n\right|_{\textrm{range}(P_W)}$, corresponds to images which only contain high frequency features with wave front vectors in $W$.

A simple implementation obtained via the $P_W$ projector which does not require changes to Curvelab can be downloaded from Github\footnote{\url{https://github.com/BolinPan/Wedge_Restricted_Curvelet}}. 
A more efficient implementation would avoid the computation of the out-of-wedge Curvelets all together but it would require making changes to Curvelab code.

%
%

\subsection{Curvelets and PAT}
\subsubsection{Curvelet representation of initial pressure}\label{sec:Curvelet:CPAT:p0}
Curvelets were shown to provide an almost optimally sparse representation if the object is smooth away from (piecewise) $\mathcal{C}^2$ singularities \cite{NC}, \cite{starck2002curvelet}, \cite{SCIOAD}. More precisely, the error of the best $s$-term approximation (corresponding to taking the $s$ largest in magnitude coefficients) in Curvelet frame of an image $g$ decays as \cite{NC}
\begin{equation}
\setlength\abovedisplayskip{3pt}
\setlength\belowdisplayskip{3pt}  
||g-g_s||^2_2 \leq \mathcal{O} ( (\log s)^3\cdot s^{-2} ). 
\end{equation}
Motivated by this result, we investigate the Curvelet representation of the initial pressure $p_0$. Here we can directly apply the symmmetrised Curvelet transform \eqref{eq:NonwrapCurveletformula} to un-mirrored $p_0$ which is a real function $\mathbb{R}^d \rightarrow \mathbb R$. In this case the frequency $\mathbf{k}$ in \eqref{eq:NonwrapCurveletformula} corresponds directly to the frequency $\mathbf{k}$ in the ambient space.
In practice, $p_0$ is discretised on an $n$ point grid in $\mathbb R^d$, $p_0 \in \mathbb R^n$, and we apply the symmetric discrete Curvelet transform $\Psi\, : \mathbb R^n \rightarrow \mathbb R^N$.

\subsubsection{Wedge restricted Curvelet representation of PAT data}\label{sec:Curvelet:RestrictedCurvelet}

The acoustic propagation is nearly optimally sparse in the Curvelet frame and so is well approximated by simply translating the Curvelet along the Hamiltonian flow \cite{dataMotivation}.
This implies the sparsity of the initial pressure in the Curvelet frame is essentially preserved by acoustic propagation. In Section \ref{sec:Curvelet:WaveFrontMapping} we derived the one to one mapping between the wavefront direction in ambient space and the corresponding direction in data domain \eqref{eq:map:kangles}. We can use this relation to essentially identify a Curvelet in the ambient space (and hence image space) with the induced Curvelet in the data space, which in turn implies that the sparsity of the initial pressure is translated into the sparsity of the PAT data. Motivated by this result we consider sparse representation of the full volume of photoacoustic data in the Curvelet frame. 

The bow-tie shaped range of the PAT forward operator together with Hermitian symmetry of real functions in Fourier domain suggest the \textit{symmetrised wedge restricted Curvelet transform} introduced in \ref{sec:Curvelet:wrc} for PAT data volume representation.

For the PAT data representation, we reinterpret the notation in \eqref{eq:NonwrapCurveletformula} underlying the symmetrised wedge restricted Curvelet transform as follows
\begin{enumerate}
    \item[$\bk$] as a data domain frequency $(\omega/c, \mathbf{k_{\mathcal S}})$;
    \item[$\theta_{j,l}$] as a discrete direction $\beta_{j,l}$ of the data domain trapezoidal frequency window $\tilde{U}_{j,\beta_{j,l}}(\omega/c, \bk_{\mathcal S})$. We note that we do not use the wavefront angle map \eqref{eq:WaveFrontMapping} to compute these angles explicitly, only to argue the sparsity of the data. The tiling is computed using standard Curvelet transform applied to rectangular domain, followed by the projection on the range $P_W$ (see Sections  \ref{sec:Curvelet:CPAT:time},\ref{sec:Curvelet:CPAT:beta}). The corresponding angles $\beta_{j,l}$ are obtained via \eqref{eq:beta_l} for each quadrant;
    \item[$W$] the restriction of $\beta_{j,l}$ to those in the bow-tie $W = (-\pi/4, \pi/4)^{d-1} \cup (3/4\pi, 5/4\pi)^{d-1}$ effects the projection on the range of PAT forward operator $\mathcal R_A$ (to be precise on a set containing the range due to isotropic coarse scale functions $\tilde U_0$ but we use this term loosely throughout the paper); 
    \item[$({\bf a}_t, {\bf a}_{\mathcal S})$] the grid of translations along the normalised time axis $ct$ and the detector $\mathcal S$, respectively. Because we are using the wrapping implementation of fast digital Curvelet transform, this grid is the same for all angles within the same quadrant $\mathbb T$ at the same scale;
\end{enumerate}
This results in the following formula for wedge restricted Curvelet coefficients
\begin{align}\label{eq:Cdata}\tag{WRC}
\setlength\abovedisplayskip{3pt}
\setlength\belowdisplayskip{3pt}  
&\tilde C_{j,l}({\bf a}_t, {\bf a}_{\mathcal S}) \quad\quad\quad\quad \quad\quad\quad\quad \beta_{j,l} \in W\\ 
\nonumber &=\iint_{\mathbb R^d} \hat{u}\left(\frac{\omega}{c}, \bk_{\mathcal S}\right) \tilde{U}_{j,\beta_{j,l}}\left(\frac{\omega}{c}, \bk_{\mathcal S} \right) 
e^{i (\bx_{\bf a} \cdot (\omega/c, \bk_{\mathcal S}) )} 
d\frac{\omega}{c} d{\bk_{\mathcal S}}
\end{align}
with $\bx_{\bf a} = (2^{-j}{\bf a}_t, 2^{-j/2}{\bf a}_{\mathcal S})$.

We will demonstrate in the results section that the above restriction to the range of PAT data representation is absolutely essential to effectively eliminate artefacts associated with subsampling of the data, which align along the out of range directions.

\subsubsection{The effect of time oversampling}\label{sec:Curvelet:CPAT:time}
Any practical realisation of (wedge restricted) Curvelet transform for PAT data representation needs to account for PAT data being commonly oversampled in time w.r.t.~to the spatial resolution resulting in data volumes elongated in the time dimension. 
As the Curvelet transform is agnostic to the underlying voxel size, a.k.a.~grid spacing, to apply the wedge restriction correctly we need to calculate the speed of sound per voxel, $c_v$. To this end we first estimate $\bx_{\max}$, the longest path between any point in the domain $\Omega \in \mathbb R^d$ and the detector $\mathcal S$ (the length of the diagonal for the cuboid $\Omega$) and the corresponding longest travel time $T_{\max} = \bx_{\max}/c$. We note, that when simulating data using k-Wave, $T_{\max}$ coincides with the time over which the data is recorded. However, for a number of reasons, the real data is often measured over a shorter period of time, $T < T_{\max}$.  

Assuming an idealised data domain $(0,T_{\max}) \times \mathcal S$, we can calculate the speed of sound per voxel\footnote{$c_v$ is usually close to $0.3$ in experiments in soft tissue}, $c_v$, as a ratio of the spatial grid points to temporal steps to travel $\mathbf{x}_{\max}$
\begin{equation}\label{eq:rescaling}
\setlength\abovedisplayskip{3pt}
\setlength\belowdisplayskip{3pt}  
c_v = 
\frac{\bx_{\max}/h_{\bx}}{T_{\max}/h_t} = \frac{c h_t }{h_\bx}.
\end{equation}
Here $h_{\bx}$ and $h_t$ are the sensor/spatial grid spacing (we assume the same grid spacing in each dimension in space and coinciding with the grid at the sensor), and the temporal step, respectively. Plausibly, $c_v$ is independent of the number of time steps in the data i.e.~it is the same for any measurement time $T$. 

So far, we have been consistently using normalised time i.e.~$ct$ instead of $t$ (frequency $\omega/c$ instead of $\omega$) which corresponds to the range of PAT forward operator $\beta \in \left(-\frac{\pi}{4}, \frac{\pi}{4} \right)^{d-1} \cup \left(\frac{3\pi}{4}, \frac{5\pi}{4} \right)^{d-1}$. 
To illustrate the effect of time oversampling we will now shortly resort to the non-normalised formulation. This is for illustrative purpose only and exclusively the normalised formulation is used in computations i.e.~the angle $\beta$ always refers to the normalised formulation.

\begin{figure}[h!]
\vspace{-0.7cm} 
\setlength{\abovecaptionskip}{0.1cm} 
\setlength{\belowcaptionskip}{-1cm}   
  \subfloat[]{
  \includegraphics[width=0.23\linewidth,height=0.23\linewidth]{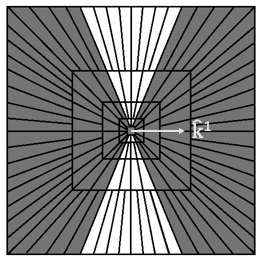}
}
\hfill
  \subfloat[]{
  \includegraphics[width=0.46\linewidth,height=0.23\linewidth]{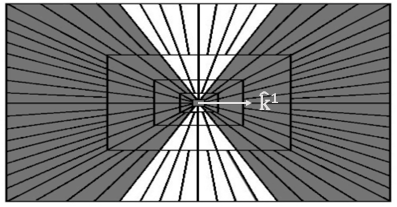}
  }
  \hfill
  \subfloat[]{\includegraphics[width=0.23\linewidth,height=0.23\linewidth]{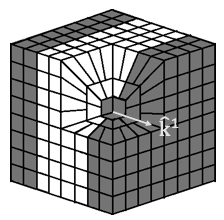}}
  \caption[2Dgrey]{An illustration of the bow-tie range of the PAT forward operator mapped over a 2D Curvelet tilling in (a) non-normalised data frequency space $(\omega, \bk_\mathcal{S})$ (note asymptotes $\pm\omega/c_v$), (b) normalised data frequency space $(\omega/c_v, \bk_\mathcal{S})$ restricted to the sensor frequency range (this form underlies the discrete Curvelet transform \cite{3DFDCT}). The Curvelets in the range are depicted in grey and these out of range in white. We stress that due to our definition of the angle $\beta$ the corresponding wavefront vector in our coordinate system is $\bar\beta = (\cos\beta, -\sin\beta)$ which sweeps the grey region while the angle $\beta$ sweeps $ (-\pi/4,\pi/4) \cup (3/4\pi, 5/4 \pi)$; (c) 3D equivalent of (a). Visualisation corresponds to $c_v = 0.5$.}  
  \label{image:2Dgrey}
\end{figure}

In the non-normalised formulation the speed of sound per voxel $c_v$ determines the slope of the asymptotes in the voxelized domain, $\pm c_v t$. 
Taking the Fourier transform in each dimension, we obtain the corresponding asymptotes in frequency domain as $\pm \omega / c_v$.
Figure \ref{image:2Dgrey} illustrates the effect of the time oversampling i.e.~$c_v < 1$ on the Curvelet discretized range of the PAT forward operator in two and three dimensions. 

\dontshow{
\begin{figure}[h!]
  \centering
  \includegraphics[width=0.5\linewidth,height=0.5\linewidth]{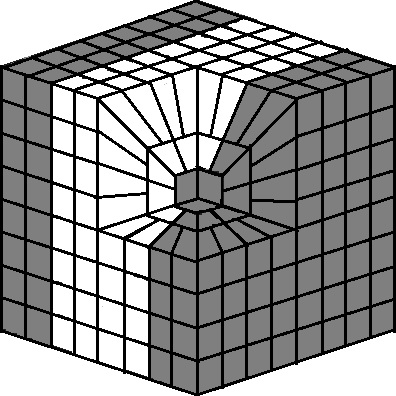}
  \caption[3Dgrey]{An illustration of the bow-tie range of the PAT forward operator mapped over a 3D Curvelet tilling in the frequency domain. The Curvelets in the range are depicted in grey and these out of range in white.}
  \label{image:3Dgrey}
\end{figure}}

\subsubsection{Discretisation of the orientation $\beta_{j,l}$}\label{sec:Curvelet:CPAT:beta}
Figure \ref{image:2Dgrey}(b) depicts the tiling underpinning the discrete Curvelet transform on a rectangular domain \cite{FDCT}. As this tiling is based on the normalised data frequency space (w.r.t.~the sound speed per voxel $c_v$), 
the slopes of the asymptotes remain $\pm 1$ and the range of the PAT forward operator remains unaltered in this representation, $\beta \in (-\pi/4,\pi/4)^{d-1} \cup (3/4\pi, 5/4 \pi)^{d-1}$. 

Taking standard Curvelet transform as defined in \cite{FDCT} of time-oversampled discretised data volume corresponds to discretising the normalised data frequency domain $(\omega/c_v, {\bf k}_{\mathcal S})$ with maximal frequency $|\omega^{\max}/c_v| = 1/(T c_v)$ along time and $|{\bf k}_{\mathcal S}^{\max, i}|, \, i=1,\dots, d-1$ along the detector coordinates.
In effect, the highest represented frequency in the discretised data volume along the time axis $|\omega^{\max}/c_v| = 1/(T c_v)$  is $1/c_v$ times  higher than the corresponding highest frequency along the sensor coordinates $|{\bf k}_{\mathcal S}^{\max, i}|, \, i=1,\dots, d-1$. This affects the sampling of the discrete angles $\beta_{j,l}$ because it changes the ratio of the sides of the cuboid domain.

The wedge orientations $\beta_{j,l}$ in the discrete Curvelet transform \cite{FDCT} are obtained via equisampling the tangents of their corresponding angles $\hat \beta_{j,l}^{\mathbb T}$. In 2D for each of the four quadrants (in 3D equivalently 6 pyramids \cite{3DFDCT}): north, west, south and east, defined by the corners and the middle point of the rectangular domain, for each wedge angle $\beta_{j,l}$, $\hat \beta_{j,l}^{\mathbb T}$ denotes its version measured from the central axis of the corresponding quadrant $\mathbb T = \{\mathbb  N, \mathbb W, \mathbb S, \mathbb  E\}$. Consequently, $\tan \hat \beta_{j,l}^{\mathbb N,\mathbb S}$ in the north/south quadrant are scaled with $1/c_v$ while $\tan \hat \beta_{j,l}^{\mathbb W, \mathbb E}$ in the west/east triangle are scaled with $c_v$  
\begin{align}\label{eq:beta_l}
\setlength\abovedisplayskip{3pt}
\setlength\belowdisplayskip{3pt}  
          \hat \beta_{j,l}^{\mathbb N, \mathbb S} = \arctan \frac{1}{c_v} \frac{l-1/2}{L_j/8},  \quad l=-L_j/8+1,..,L_j/8, \\
\nonumber      \hat \beta_{j,l}^{\mathbb W, \mathbb E} = \arctan c_v \frac{l-1/2}{L_j/8}, \quad l=-L_j/8+1,..,L_j/8,
\end{align} 
where $L_j = L\cdot 2^{\lfloor j/2 \rfloor}$ is the total number of angles at the scale $j$ which doubles every second scale. Note that the slope of the boundary between two neighboring quadrants is not $\pm 1$ but $\tan 1/c_v$ or $\tan c_v$, respectively.
The side effect of this scaling is that the sampling of the angles in the range is more dense than the angles out of range (one in four angles $\beta_{j,l}$ are out of the range at each scale in Figure \ref{image:2Dgrey}), which plays in favour of our method.

%
\section{Optimization}\label{sec:Optimization}
%
%
Let $\Phi \in \mathbb{R}^{m\times n}$ be the sensing matrix the rows of which are the respective sensing vectors $\phi_i\tr, i=1,\dots,m$ with $m\ll n$ and $\Psi:\mathbb{R}^n\to\mathbb{R}^N$ be the isometric sparsifying transform with $\Psi^{\dagger} = \Psi\tr$ its left inverse. 
We model the compressed measurements $b\in \mathbb{R}^m$ as
\begin{equation}
\setlength\abovedisplayskip{3pt}
\setlength\belowdisplayskip{3pt}
b = \Phi g + e = \Phi \Psi^{\dagger} f + e,
\label{eq:CSmodel}
\end{equation}
where $e$  is the measurement noise (with $\|e\|_2 \leq \varepsilon$ for some small $\varepsilon$), and $f$ is the sparse representation of the original signal $g \in \mathbb{R}^n$ in the basis $\Psi$, \emph{i.e.}~$f=\Psi g$. Under conditions specified in \cite{stableRecovery,RIPrelaxe}, 
the sparse signal $f$ 
can be recovered from the compressed measurements $b$ via the basis pursuit ($\ell_1$ minimization problem) 
\begin{equation}\label{eq:ell1Minimization}
\setlength\abovedisplayskip{3pt}
\setlength\belowdisplayskip{3pt}  
\min_{f\in \mathbb{R^N}} \|f\|_1 \quad \text{s.t.}  \; \|\Phi \Psi^{\dagger} f  - b \|_2 \leq \varepsilon.
\end{equation}
Furthermore, an (approximately) $s$-sparse signal $f$ can be robustly recovered with overwhelming probability if $m\geq C\cdot\log(n)\cdot s$ for some positive constant $C$ \cite{Robust,stableRecovery} .

\subsection{Iteratively reweighted $\ell_1$ norm}\label{sec:Optimization:IRL1}

Efficient representation in an overdetermined basis requires highly sparse solutions.
In contrast to $\ell_0$ semi-norm where each non-zero coefficient is counted regardless of its magnitude, the $\ell_1$ penalty is proportional to the coefficient's magnitude resulting in small coefficients being penalised less than the large coefficients and consequently in less sparse solutions. 
We address this issue using weighted formulation of $\ell_1$ norm penalty, $||\Lambda f||_1$, 
where $\Lambda$ is a diagonal matrix with positive weights and $f$ is the sparse representation of the solution in a basis $\Psi$.
Reweighted $\ell_1$ penalty was shown in many situations to outperform the standard $\ell_1$ norm in the sense that substantially fewer measurements are needed for exact recovery \cite{enhancing}, \cite{IRLSMSR}, \cite{SSLS}, \cite{UMSS}. 

To enhance the sparsity of the solution, the weights can be chosen as inversely proportional to the magnitudes of the components of the true solution. As the latter is not available, the current approximation $f^k$ is used in lieu \cite{enhancing} 
\begin{equation}\label{eq:lambda}
\setlength\abovedisplayskip{3pt}
\setlength\belowdisplayskip{3pt}  
\Lambda^{k} = \diag{1/(|f^k|+\epsilon)}.
\end{equation}
Initialisation choosing $\Lambda^1 = \mbox{diag}({\bf 1})$ i.e.~with the non-weighted $\ell_1$ seems to work well in practice.
We follow an adaptive scheme for the choice of $\epsilon$ proposed in \cite{enhancing} (another approach is suggested in \cite{IRLSMSR}). 
In each iteration $k$, we first normalize the magnitude of $f^k$, denoted $\bar{f}^k$. 
Let $\rho(\bar{f}^k)$ be the non-increasing arrangement of $\bar{f}^k$, and $\rho(\bar{f}^k)_S$ be the $S$-th element after reordering.
Then we update $\epsilon$ as
\begin{equation}
\setlength\abovedisplayskip{3pt}
\setlength\belowdisplayskip{3pt}  
\epsilon = \max \{\rho(\bar{f}^k)_S,10^{-4}\}.
\end{equation}
Such choice of $\epsilon$ is geared towards promoting $S$-sparse solutions.
As Curvelet frame is heavily overdetermined, we propose to use reweighted $\ell_1$ for both data reconstruction and $p_0$ reconstruction. Recall the $s$-sparse signals can be recovered exactly if $m \geq C\cdot \log(n)\cdot s$, and so the sparsity $S$ is chosen to be as $m/(C\cdot \log(n))$ for some positive constant $C$. A version of $\ell_1$ reweighting used in our later algorithms is summarized in Algorithm ${\bf W}$.
\begin{W}
\renewcommand{\thealgorithm}{}
\small 
\caption{$\Lambda \leftarrow \mbox{weight}(f, S)$  \cite{enhancing}}
\label{algo:R}
\begin{algorithmic}[1]
\STATE \textbf{Input:} $f$, $S$ \par
\STATE \textbf{Output:} $\Lambda$ \par
\STATE $\bar{f} = |f|/\max(|f|)$\par
\STATE $\epsilon = \max(\rho(\bar{f})_S,10^{-4})$\par
\STATE $\Lambda = \diag{1/(|f|+\epsilon)}$\par
\end{algorithmic}
\end{W}

%
%
\subsection{Data reconstruction}\label{sec:Optimization:DR}
In this two step approach we propose to first reconstruct the full volume photoacoustic data from subsampled data as described below, followed by the recovery of initial pressure (photoacoustic image) using time reversal via a first order method implemented in the k-Wave Toolbox \cite{kWave}.\par
We formulate the data reconstruction problem as an iteratively reweighted $\ell_1$ minimisation problem
\begin{equation}\label{DR}\tag{\bf DR}
\setlength\abovedisplayskip{3pt}
\setlength\belowdisplayskip{3pt}  
f^{k+1} \xleftarrow{k} \argminA_f \frac{1}{2}||\Phi {\tilde{\Psi}}^{\dagger}f-b||_2^2 + \tau ||\Lambda^k f ||_1,
\end{equation}
where $\Phi \in \mathbb R^{m\times n_b}$ is the subsampling matrix, $b \in \mathbb{R}^{n_b}$ is the subsampled PAT data, ${\tilde{\Psi}}^{\dagger}: \mathbb R^{\tilde N_b} \rightarrow \mathbb R^{n_b}$ is the left inverse (and adjoint) of the wedge restricted Curvelet transform of the PAT data $\tilde\Psi$ (see \ref{sec:Curvelet:wrc}, \ref{sec:Curvelet:RestrictedCurvelet}, \ref{sec:Curvelet:CPAT:time}, \ref{sec:Curvelet:CPAT:beta}), $f$ are the coefficients of the full PAT data in $\tilde \Psi$ 
and $\Lambda^k$ is a diagonal matrix of weights updated at each iteration using procedure ${\bf W}$.\par

\begin{R-SALSA}
\renewcommand{\thealgorithm}{}
\small 
\caption{Reweighted Split Augmented Lagrangian Shrinkage Algorithm \cite{SALSA}}
\begin{algorithmic}[1]
\STATE \textbf{Initialization:} $y^1 = \mathbf{0}$, $w^1 = \mathbf{0}$, $f^1 = \mathbf{1}$, $\Lambda^1 = \diag{\mathbf{1}}$, $\mu>0$, $\tau>0$, $\eta>0$ and $K_{\text{max}}$\par
\STATE $k := 1$\par
\STATE \textbf{Repeat}\par
\STATE $f^{k+1} = \argminA_f ||\Phi {\tilde{\Psi}}^{\dagger} f - b||_2^2 + \mu ||f - y^k-w^k||_2^2$\par
\STATE $y^{k+1} = \argminA_y \tau||\Lambda^k y||_1+\frac{\mu}{2}||f^{k+1}-y-w^{k}||_2^2$\par
\STATE $w^{k+1} = w^k -(f^{k+1} - y^{k+1})$\par
\STATE $\Lambda^{k+1} = \mbox{weight}(f^{k+1}, S)$\par
\STATE $k = k+1$\par
\STATE \textbf{Until} $||f^{k}-f^{k-1}||_2 /||f^{k-1}||_2<\eta \text{ or } k>K_{\text{max}}$
\end{algorithmic}
\label{algo:DR}
\end{R-SALSA}

At each iteration of the iteratively reweighted $\ell_1$ scheme for \ref{DR}, the minimisation problem for a fixed $\Lambda^k$ is solved with an ADMM type scheme, SALSA \cite{SALSA}, resulting in a reweighted version of SALSA summarized in Algorithm \textbf{R-SALSA}. The sub-problem in \textbf{R-SALSA} line 4 is a quadratic minimization problem; its solution can be obtained explicitly via the normal equations
\begin{equation}\label{eq:DRLS}
\begin{aligned}
\setlength\abovedisplayskip{3pt}
\setlength\belowdisplayskip{3pt}  
f^{k+1} & = (\tilde{\Psi}\Phi^{\rm T}\Phi {\tilde{\Psi}}^{\dagger} + \mu I)^{-1}(\tilde{\Psi} \Phi^{\rm T}b + \mu(y^k+w^k))
\end{aligned}
\end{equation}
and using the Sherman--Morrison--Woodbury formula, the inverse in \eqref{eq:DRLS} can be reformulated as
\begin{equation}
\setlength\abovedisplayskip{3pt}
\setlength\belowdisplayskip{3pt}  
(\tilde{\Psi}\Phi^{\rm T}\Phi{\tilde{\Psi}}^{\dagger}+\mu I)^{-1} = \frac{1}{\mu}\left(I - \frac{1}{\mu+1}\tilde{\Psi}\Phi^{\rm T}\Phi{\tilde{\Psi}}^{\dagger}\right).
\end{equation}
The solution of the proximal w.r.t.~$\frac{\tau}{\mu} \Lambda^k \| \cdot \|_1$ sub-problem in line 5 of \textbf{R-SALSA} amounts to applying to $\tilde{y}^k = f^{k+1}-w^k$ the following soft thresholding operator with vector valued threshold $S_{\frac{\tau}{\mu}\Lambda^k}$ 
\begin{equation}
\setlength\abovedisplayskip{3pt}
\setlength\belowdisplayskip{3pt}  
S_{\frac{\tau}{\mu}\Lambda^k}(\tilde{y}^k) = \textrm{sign}\left(\tilde{y}^k \right) \odot \left(|\tilde{y}^k| - \frac{\tau}{\mu}\Lambda^k \mathbf{1}\right),
\end{equation}
where $|\cdot|$ is taken pointwise and $\odot$ denotes pointwise multiplication.

%
%
\subsection{$p_0$ reconstruction}\label{sec:Optimization:p0R}
In this one step approach we reconstruct the initial pressure $p_0$ (photoacoustic image) directly from subsampled photoacoustic data. We formulate the $p_0$ reconstruction 
as an iteratively reweighted $\ell_1$ minimization problem 
\begin{equation} \label{p0R} \tag{\bf $p_0$R}
\setlength\abovedisplayskip{3pt}
\setlength\belowdisplayskip{3pt}  
f^{k+1} \xleftarrow{k} \argminA_f \frac{1}{2}||\Phi A \Psi^{\dagger}f-b||_2^2 + \tau ||\Lambda^k f||_1,
\end{equation}
where $\Phi \in \mathbb{R}^{m\times n_b}$ and $b \in \mathbb{R}^{n_b}$ are, as before, the subsampling matrix and the corresponding subsampled PAT data, $A : \mathbb{R}^{n_{p_0}} \rightarrow  \mathbb{R}^{n_b}$ is the 
photoacoustic forward operator, $\Psi^{\dagger} : \mathbb{R}^{N_{p_0}} \rightarrow  \mathbb{R}^{n_{p_0}} $ is the left inverse (and adjoint)
of the standard Curvelet transform $\Psi$ (see \ref{sec:Curvelet:CurveletTransform}, \ref{sec:Curvelet:CPAT:p0}), $f$ are the coefficients of $p_0$ in $\Psi$ and $\Lambda^k$ is a diagonal matrix of weights updated at each iteration using procedure ${\bf W}$.\par

\begin{R-FISTA}
\renewcommand{\thealgorithm}{}
\small 
\caption{Reweighted Fast Iterative Shrinkage Thresholding Algorithm \cite{FISTA}}
\begin{algorithmic}[1]
\STATE \textbf{Initialization:} $y^1 = \mathbf{0}$, $f^1 = \mathbf{1}$, $\Lambda^1 = \diag{\mathbf{1}}$, $\alpha^1 = 1$, $\mu=\frac{1}{L}$, $\tau>0$, $\eta>0$ and $K_{\text{max}}$ \par
\STATE $k := 1$\par
\STATE \textbf{Repeat}\par
\STATE $\tilde{z}^k = y^k - \mu \Psi A^{\rm T}\Phi^{\rm T}(\Phi A \Psi^{\dagger}y^k - b)$\par
\STATE $f^{k+1} = \argminA_f \mu\tau||\Lambda^k f||_1+\frac{1}{2}||f-\tilde{z}^k||_2^2$\par
\STATE $\alpha^{k+1} = (1+\sqrt{1+4(\alpha^k)^2})/2$\par
\STATE $y^{k+1} = f^{k+1} + \frac{\alpha^k-1}{\alpha^{k+1}}(f^{k+1} - f^{k})$\par
\STATE $\Lambda^{k+1} = \mbox{weight}(f^{k+1}, S)$\par
\STATE $k = k+1$ \par
\STATE \textbf{Until} $||f^{k}-f^{k-1}||_2 /||f^{k-1}||_2<\eta \text{ or } k>K_{\text{max}}$
\end{algorithmic}
\label{algo:p0R}
\end{R-FISTA}

Due to a high cost of application of the PAT forward $A$ and adjoint $A^{\rm T}$ operators, we propose to solve \ref{p0R} with FISTA \cite{FISTA} which avoids inner iteration with $A$ and $A^{\rm T}$.  
At each iteration of FISTA we solve the proximal $\Lambda^k$ weighted $\ell_1$ problem and subsequently update the weights as detailed in Algorithm \textbf{R-FISTA} (lines 5, 8). In the \textbf{R-FISTA} algorithm, $L$ is an approximation to $\| \Psi A^{\rm T}\Phi^{\rm T}\Phi A \Psi^{\dagger} \|_2$,
which can be pre-computed for given $A, \Psi$ and a subsampling scheme $\Phi$, using a simple power iteration.

%
%
\subsection{$p_0$ reconstruction with non-negativity constraint}\label{sec:Optimization:p0R+}
Considering the non-negativity constraint as additional prior information on photoacoustic image, we formulate the non-negativity constraint version of  $p_0$ reconstruction 
\begin{equation}\label{p0R+}\tag{$p_0${\bf R+}}
\begin{aligned}
\setlength\abovedisplayskip{3pt}
\setlength\belowdisplayskip{3pt}  
f^{k+1} \xleftarrow{k} & \argminA_{f} \frac{1}{2}||\Phi A \Psi^{\dagger}f-b||_2^2 + \tau ||\Lambda^k f||_1, \\
& \text{ s.t. } \Psi^{\dagger} f \geq 0.
\end{aligned}
\end{equation}
To incorporate the non-negativity constraint into the objective function we rewrite \ref{p0R+} with $p = \Psi^{\dagger}f$
\begin{equation}\label{eq:p0R+Refomulated}
\begin{aligned}
\setlength\abovedisplayskip{3pt}
\setlength\belowdisplayskip{3pt}  
p^{k+1} \xleftarrow{k} & \argminA_{p\geq0} \frac{1}{2}||\Phi A p-b||_2^2 + \tau ||\Lambda^k \Psi p||_1\\
&= \argminA_{p} \frac{1}{2}||\Phi A p-b||_2^2 + \tau ||\Lambda^k \Psi p||_1+ \chi_{+}(p),
\end{aligned}
\end{equation}
where $\chi_{+}(p)$ is the characteristic function of non-negative reals applied pointwise.
The non-negativity constraint $p_0$ reconstruction in a form \eqref{eq:p0R+Refomulated} is amiable to solution with ADMM \cite{ADMM} inside an iteratively reweighted $\ell_1$ minimization scheme. We first introduce an auxiliary variable
\begin{equation}
\setlength\abovedisplayskip{3pt}
\setlength\belowdisplayskip{3pt}  
y 
= \begin{pmatrix}
y_1 \\ y_2
\end{pmatrix}
= \begin{pmatrix}
\Psi \\ I_{n_{p_{0}}}
\end{pmatrix}p = Ep.
\end{equation}
Then \eqref{eq:p0R+Refomulated} is equivalent to
\begin{equation}
\begin{aligned}
\setlength\abovedisplayskip{3pt}
\setlength\belowdisplayskip{3pt}  
\argminA_{p, y} \text{ } &G(p)+H(y) \quad \text{s.t.} \; y = Ep,
\end{aligned}
\end{equation}
where the functions $G$ and $H$ are
\begin{equation}
\begin{aligned}
\setlength\abovedisplayskip{3pt}
\setlength\belowdisplayskip{3pt}  
G(p) & = \frac{1}{2}||\Phi A p - b||_2^2,\\
H(y) & = \tau||\Lambda^k y_1||_1+\chi_{+}(y_2).
\end{aligned}
\end{equation}
The corresponding dual variable $w = {(w_1, w_2)}^{\rm T}$ follows an analogous split. A reweighted version of ADMM used in this work is summarized in Algorithm \textbf{R-ADMM}.\par

Taking advantage of the isometry of the Curvelet transform $\Psi$, the update in line 4 of Algorithm \textbf{R-ADMM} can be reformulated as a solution of a large scaled least-squares problem
\dontshow{
\begin{equation}
\begin{aligned}
\setlength\abovedisplayskip{3pt}
\setlength\belowdisplayskip{3pt}  
p^{k+1} & = \argminA_p 
\left\{
\frac{1}{2}\left\|
\begin{pmatrix}
\Phi A \\ \sqrt{\mu}\Psi \\ \sqrt{\mu} I_{n_{p_{0}}}
\end{pmatrix}
p - 
\begin{pmatrix}
b \\ \sqrt{\mu}(y_1^k - w_1^k)\\ \sqrt{\mu}(y_2^k - w_2^k)
\end{pmatrix}\right\|_2^2
\right\} \\
& = (A^{\rm T}\Phi^{\rm T}\Phi A + 2\mu I_{n_{p_{0}}} p)^{-1}(A \Phi^{\rm T}b + \mu \Psi^{\dagger}(y_1^k-w_1^k)\\
& \quad + \mu(y_2^k-w_2^k)),
\end{aligned}
\end{equation}
}
which is solved using the \emph{Conjugate Gradient Least Square} (\emph{CGLS}) method \cite{CGLS}. The update in line 5 of Algorithm \textbf{R-ADMM} can be further split into two sub-problems, each of which admits an analytical solution
\begin{equation}
\begin{aligned}
\setlength\abovedisplayskip{3pt}
\setlength\belowdisplayskip{3pt}  
y_1^{k+1} & = \argminA_{y_1} \frac{1}{2}||\Psi p^{k+1} - y_1^k +w_1^k||_2^2 + \frac{\tau}{\mu} ||\Lambda^k y_1^k||_1\\
& = S_{\frac{\tau}{\rho}\Lambda^k} (\Psi p^{k+1} + w_1^k),\\
y_2^{k+1} & = \argminA_{y_2} \frac{\rho}{2}||p^{k+1}-y_2^k+w_2^k||_2^2 + \chi_{+}(y_2^k)\\
& = \max(0, p^{k+1} + w^k_2).
\end{aligned}
\end{equation}
\begin{R-ADMM}
\renewcommand{\thealgorithm}{}
\small 
\caption{Reweighted Alternating Direction Method of Multipliers Algorithm \cite{ADMM}}
\begin{algorithmic}[1]
\STATE \textbf{Initialization:} $y^1 = \mathbf{0}$, $w^1 = \mathbf{0}$, $f^1 = \mathbf{1}$, $\Lambda^1 = \diag{\mathbf{1}}$, $\mu>0$, $\tau>0$, $\eta>0$ and $K_{\text{max}}$\par
\STATE $k := 1$\par
\STATE \textbf{Repeat}\par
\STATE $p^{k+1} = \argminA_p G(p) + \frac{\mu}{2}||Ep-y^k+w^k||_2^2$
\STATE $y^{k+1} = \argminA_y H(y) + \frac{\mu}{2}||Ep^{k+1}-y+w^k||_2^2$ 
\STATE $w^{k+1} = w^k + E p^{k+1} - y^{k+1}$\par
\STATE $f^{k+1} = \Psi p^{k+1}$\par
\STATE $\Lambda^{k+1} = \mbox{weight}(f^{k+1}, S)$\par
\STATE $k=k+1$\par
\STATE \textbf{Until} $||f^{k}-f^{k-1}||_2 /||f^{k-1}||_2<\eta \text{ or } k>K_{\text{max}}$
\end{algorithmic}
\label{algo:p0R+}
\end{R-ADMM}

%
%
\section{Photoacoustic Reconstruction}\label{sec:PATRec}
We present our results of the data reconstruction \ref{DR} and $p_0$ reconstructions \ref{p0R}, \ref{p0R+}, using 25\% randomly subsampled measurements on a simulated data and a real 3D Fabry-P\'erot raster scan data. The time reversal \ref{eq:TRequation} and the PAT forward \ref{eq:PATforwardequation} and adjoint \ref{eq:adjoint} operators are implemented with the k-Wave Toolbox \cite{kWave}. If non-negativity of initial pressure is not explicitly enforced by the algorithm, negative values are clipped to 0 in a post-processing step.

%
%
\subsection{2D vessel phantom: rigorous comparison}\label{sec:PATRec:Phantom}
\begin{figure}[htbp!]
\centering
\vspace{-0.4cm} 
\setlength{\abovecaptionskip}{0.1cm} 
\setlength{\belowcaptionskip}{-1cm}   
  \subfloat{
  \includegraphics[width=0.5\linewidth,height=0.15\linewidth]{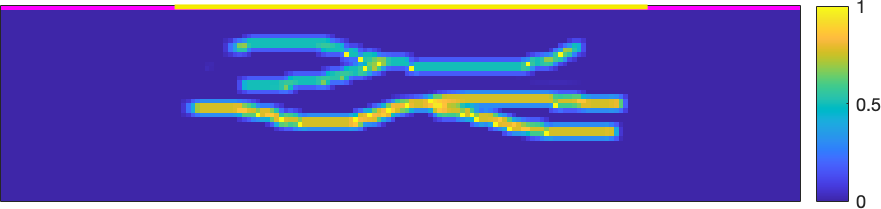}}
\caption{Vessel phantom. Position of the line sensor is marked in purple and a window with higher probability of being sampled is highlighted in yellow.}
  \label{image:p0Vessel}
\end{figure}

In an attempt to rigorously compare the two approaches \ref{DR} and \ref{p0R} on a realistic scenario we created a 2D vessel phantom in Figure~\ref{image:p0Vessel}.
This phantom is a 2D idealisation of a typical flat panel detector geometry setup where the depth of the image, $n_{\bx_\perp}$, is smaller comparing to the sensor size ($\sim n_{\mathcal S}^{1/(d-1)}$, where $n_{\mathcal S}$ is the cardinality of the regular grid on the sensor), resulting in an image resolution of $n_{\bx_\perp} \times n_{\mathcal S} = 42\times172$ with uniform voxel size $h_\bx=11.628\upmu\text{m}$.
We assume homogeneous speed of sound $c$ = 1500m/s and medium density 1000kg/$\text{m}^3$. The pressure time series recorded every $h_t$ = 29.07ns by a line sensor on top of the object with image matching resolution, i.e.~$h_{\bx}=11.628\upmu\text{m}$, constitutes the PAT data which we endow with additive white noise with standard deviation $\sigma = 0.01$. We henceforth interpret this data as a time-steps $\times$ sensors shaped volume.

\subsubsection{Fair comparison}
We note that the PAT data volume is always larger than the original initial pressure for two reasons: the measurement time $T_{\max}$ needs to be long enough for signals to arrive from the farthest corner of the domain (using notation in Section \ref{sec:Curvelet:CPAT:time}, the diagonal of the cube $\Omega$ in voxels is $\lceil \bx_{\max}/h_\bx \rceil$) and it is a common practice to oversample the PAT data in time. 
Thus to compare \ref{DR} and \ref{p0R}, \ref{p0R+} approaches in a realistic scenario fairly we upscale the reconstructed initial pressure as follows. 
We recall the definition of $c_v$ in Section \ref{sec:Curvelet:CPAT:time}, $c_v =  (\bx_{\max}/h_\bx) /  (T_{\max}/h_t)$, and note that $1/c_v$ is essentially the time oversampling factor resulting in the number of time sampling points $n_t = \lceil T_{\max}/h_t \rceil =  \lceil (1/c_v) \bx_{\max}/h_\bx \rceil$ and a measured PAT data volume size $n_t \times n_{\mathcal S}$.
We calculate a uniform upscaling factor $\alpha$ for the regular reconstruction grid to match the number of data points in the measured PAT data\footnote{alternatively, one could match the cardinalities of Curvelet coefficients}, $\alpha^d n_{\bx_\perp} n_{\mathcal S} = n_t n_{\mathcal S}$, yielding
$\alpha = \sqrt[d]{\frac{n_t}{n_{\bx_\perp}}}.$
An alternative approach would be to downscale the data during its reconstruction by representing it in a low frequency Curvelet frame introduced in \cite{Acoustic} which also allows the wedge restriction.

For the vessel phantom we choose $c_v = 0.3$ resulting in PAT volume size $591 \times 172$ and leading to $\alpha\approx 3.75$ and reconstruction grid $158\times645$ with spacing $h_\bx \approx 3.1\upmu\text{m}$. For each of these point matched grids we construct a Curvelet transform with the same number of  scales and angles to as close as possible mimic the one to one wavefront mapping. 
For the reconstructed $p_0$ image of size $158 \times 645$ we use a Curvelet transform with 4 scales and 128 angles (at the 2nd coarsest level).
For the data grid $n_t \times n_{\mathcal S} = 591 \times 172$, we start with a standard Curvelet transform with 4 scales and 152 angles (at the 2nd coarsest level). This choice, after removing the out of range angles $\beta_l \not\in (-\pi/4,\pi/4) \cup (3/4 \pi,5/4\pi)$, results in a wedge restricted Curvelet transform with 128 angles matching the number of angles of image domain Curvelets.  

\begin{figure}[htbp!]
\vspace{-0.4cm} 
\setlength{\abovecaptionskip}{0.1cm} 
\setlength{\belowcaptionskip}{-1cm}   
\centering
  \subfloat[$g(t,\bx_{\mathcal{S}})$]{
  \includegraphics[width=0.48\linewidth,height=0.18\linewidth]{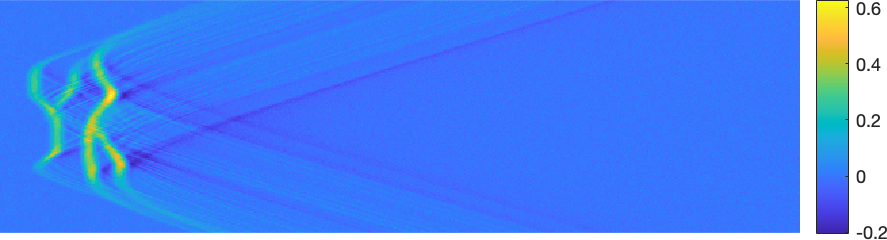}}
  \subfloat[$p_0^\textrm{TR}(g)$]{
  \includegraphics[width=0.5\linewidth,height=0.15\linewidth]{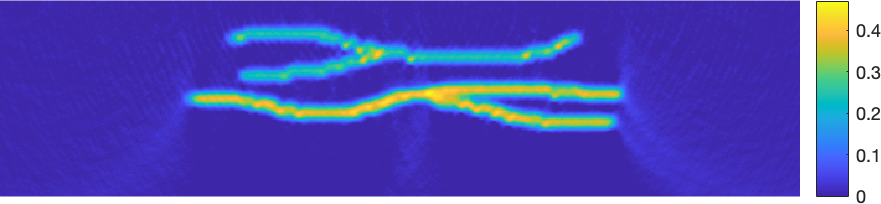}}
  \\
  \subfloat[$\hat{g}(t,\bx_{S})$]{
  \includegraphics[width=0.48\linewidth,height=0.18\linewidth]{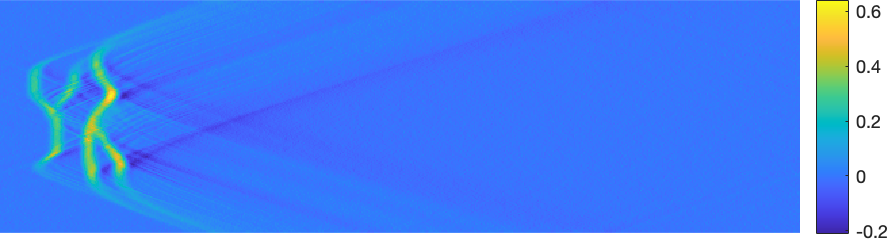}}
  \subfloat[$p_0^\textrm{TR}(\hat{g})$]{
  \includegraphics[width=0.5\linewidth,height=0.15\linewidth]{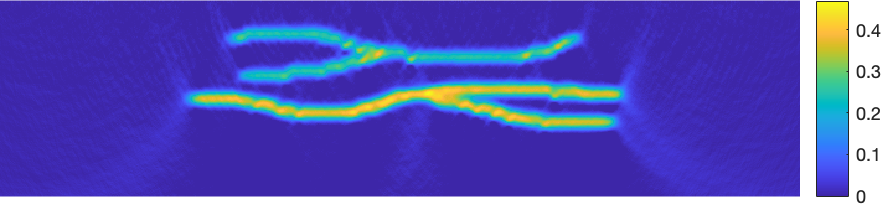}}
  \\
  \subfloat[ $\hat{g}(t,\bx_{\mathcal{S}}) -{g}(t,\bx_{\mathcal{S}})$]{
  \includegraphics[width=0.48\linewidth,height=0.18\linewidth]{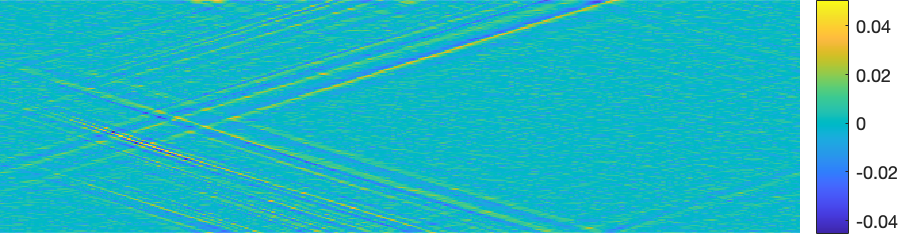}}
  \subfloat[$p_0^\textrm{TR}(\hat{g}) - p_0^\textrm{TR}(g)$]{
  \includegraphics[width=0.5\linewidth,height=0.15\linewidth]{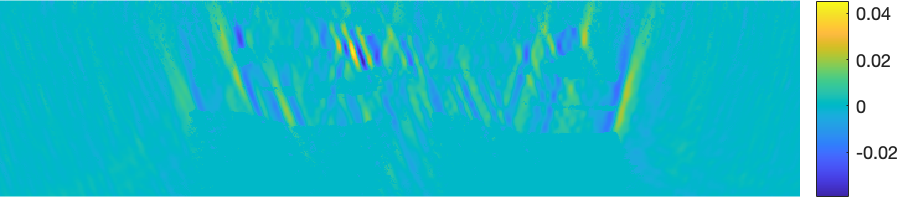}}
  \caption[VesselData]{Visualisation of accuracy of wedge restricted Curvelet transform $\tilde\Psi$ of vessel phantom data: (a) full simulated data $g(t,\bx_{\mathcal{S}})$, (b) initial pressure via \ref{eq:TRequation} of $g(t,\bx_{\mathcal{S}})$, (c) data projected on the range of $\tilde\Psi$, $\hat g(t, \bx_{\mathcal{S}}) = \tilde\Psi^\dagger \tilde\Psi g(t, \bx_{\mathcal{S}})$, (d) initial pressure via \ref{eq:TRequation} of $\hat g(t, \bx_{\mathcal{S}})$, (e) data error $\hat{g}(t, \bx_{\mathcal{S}})$ - $g(t, \bx_{\mathcal{S}})$, (f) initial pressure error $p_0^\textrm{TR}(\hat{g})$ - $p_0^\textrm{TR}(g)$.}
  \label{image:VesselData}
\end{figure}
\begin{figure}[htbp!]
\vspace{-0.6cm} 
\setlength{\abovecaptionskip}{0.1cm} 
\setlength{\belowcaptionskip}{1cm}   
\centering
  \subfloat[$|\Psi g(t, \bx_{\mathcal{S}})|$]{
  \includegraphics[width=0.32\linewidth,height=0.2\linewidth]{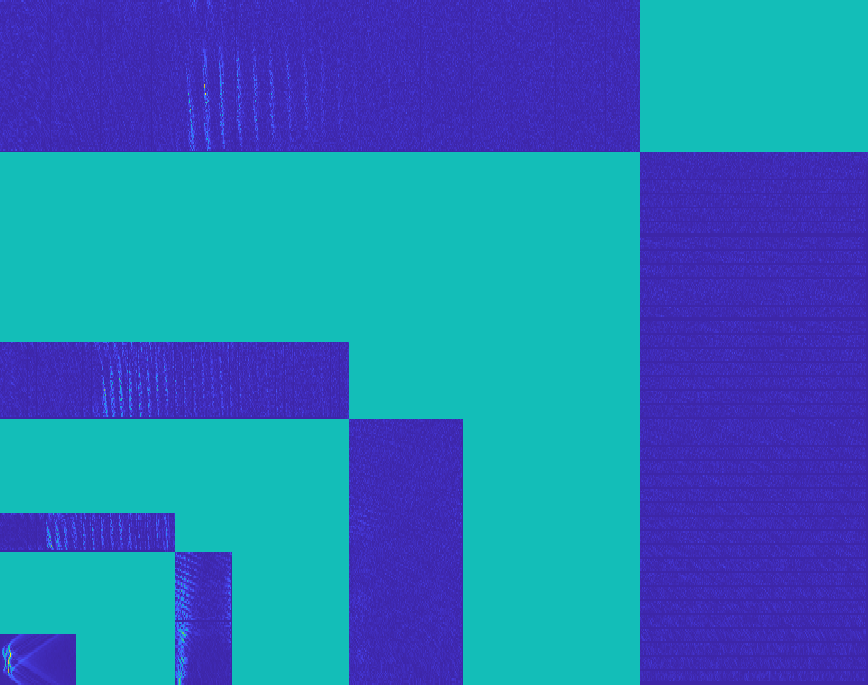}}
  \subfloat[$|\tilde{\Psi} g(t, \bx_{\mathcal{S}})|$]{
  \includegraphics[width=0.32\linewidth,height=0.2\linewidth]{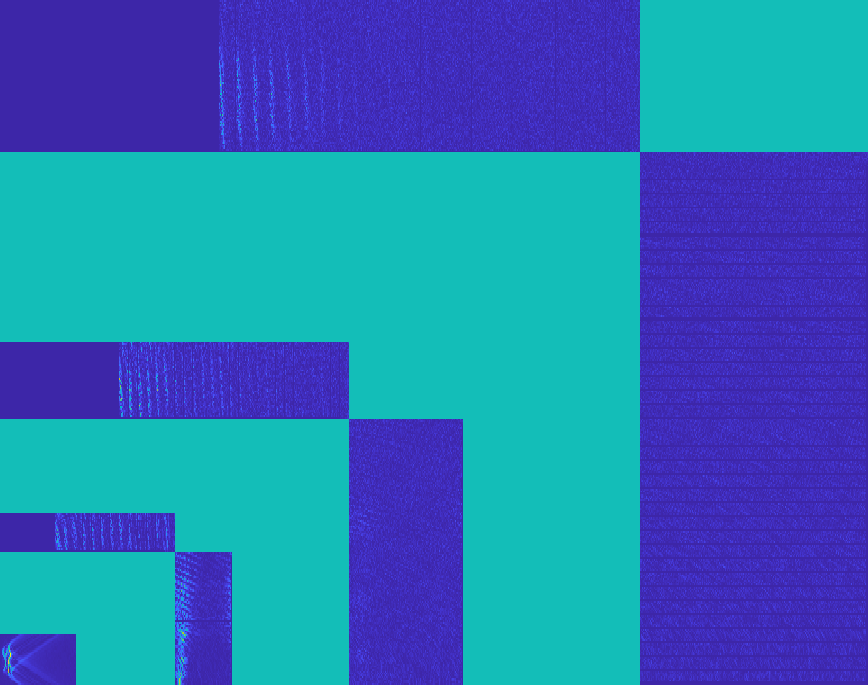}}
  \subfloat[$|\Psi p_0^\textrm{TR}(g)|$]{
  \includegraphics[width=0.32\linewidth,height=0.2\linewidth]{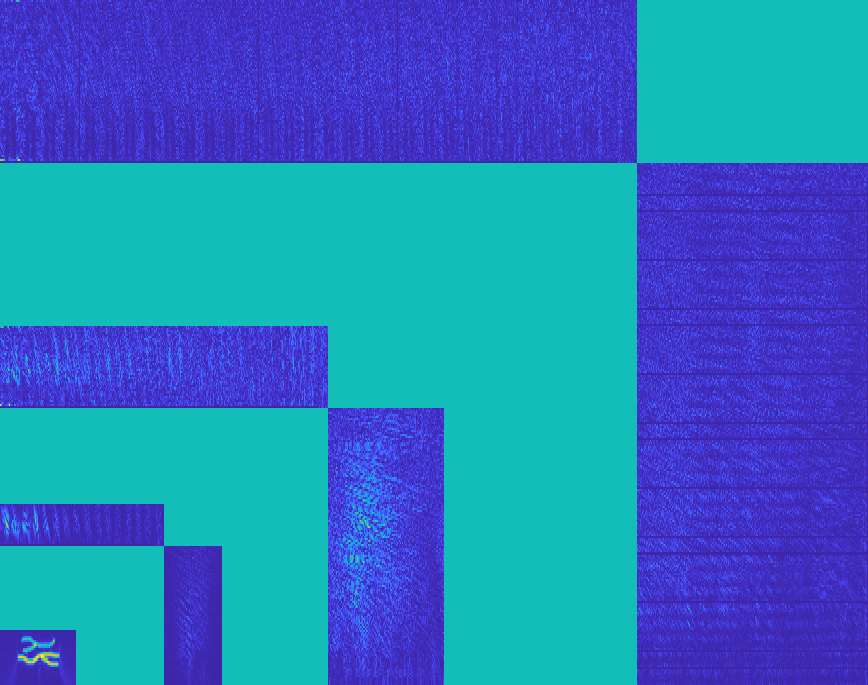}}
  \caption[VesselDataCurvelet]{Visualisation of image and data domain Curvelet coefficients (showing coarse scale overlaid over top-right quarter of higher scale scale-normalised coefficient magnitudes) of vessel phantom: (a) standard $(4,152)$-Curvelet transform of $g(t, \bx_{\mathcal{S}})$ (there are very small magnitude coefficients in the out of range wedge visible when zooming in), (b) wedge restricted $(4,128)$-Curvelet transform $g(t, \bx_{\mathcal{S}})$ (note that all coefficient out of range are 0), and (c) standard $(4,128)$-Curvelet transform of $p_0^\textrm{TR}(g)$.}
  \label{image:VesselCurvelet}
\end{figure}
\subsubsection{Representation/compression in image and data domain}
Let us first demonstrate that exclusion of the out of range angles in wedge restricted Curvelet transform $\tilde\Psi$ indeed does not impact the accuracy beyond numerical/discretisation errors. 
Figure~\ref{image:VesselData} shows the effect of wedge restricted transform $\tilde\Psi$ in data and image domain comparing (a,b) the simulated data $g(t, \bx_{\mathcal{S}})$ and its time reversal versus (c,d) the data projected on the range of $\tilde\Psi$ by applying forward and inverse wedge restricted transform $\hat g(t, \bx_{\mathcal{S}}) = \tilde\Psi^\dagger \tilde\Psi g(t, \bx_{\mathcal{S}})$ and its time reversal. The projection error is of the order of 4\% in both data (e) and image (f) domain and concentrates at the boundary of the bow-tie shaped range of PAT forward operator. In fact, (e) suggests that this is the effect of the angle discretisation with the artefacts corresponding to the wavefronts in the range of the continuous PAT forward operator but with $|\beta|$ larger than the largest discrete $|\beta_l|$ (for $\beta \in (-\pi/4,\pi/4)$ and symmetric analogy for $\beta \in (3/4 \pi/4, 5/4\pi)$). These wavefronts are mapped according to \eqref{eq:WaveFrontMapping} to image domain resulting in artefacts visible in (f). 

\begin{figure}[ht]
\vspace{-0.6cm} 
\setlength{\abovecaptionskip}{0.1cm} 
\setlength{\belowcaptionskip}{-1cm}   
  \subfloat[]{
  \includegraphics[width=0.48\linewidth,height=0.3\linewidth]{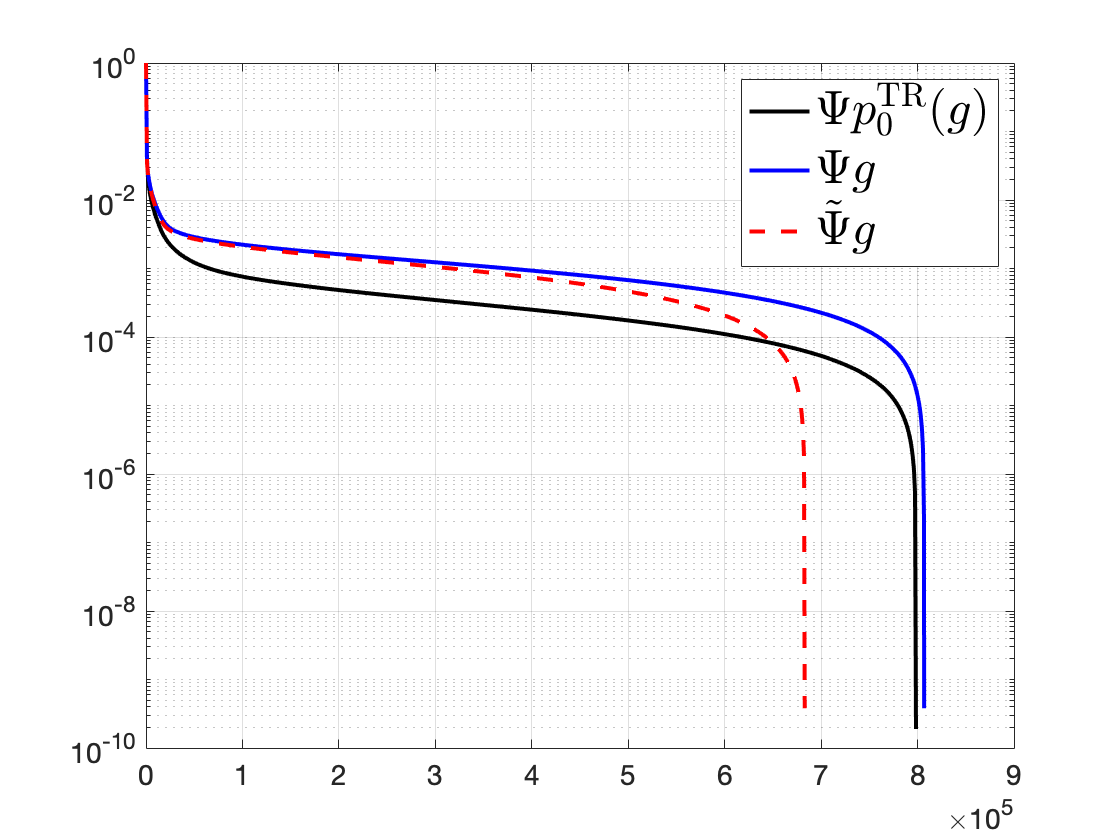}}\hfill
  \subfloat[]{
  \includegraphics[width=0.48\linewidth,height=0.3\linewidth]{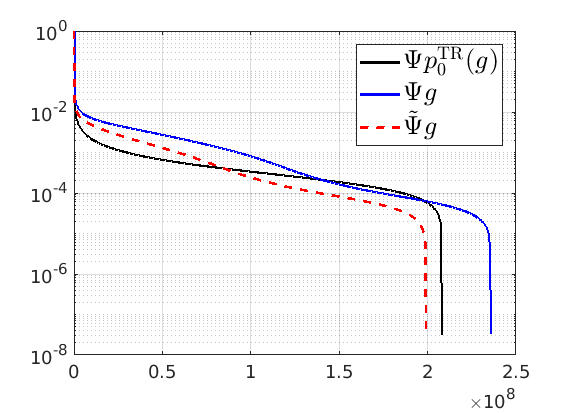}} 
  \caption[CoeffLog]{The decay of log amplitudes of image and data Curvelet coefficients $\Psi p_0^\textrm{TR}(g)$, $\Psi g(\mathbf{x}_{\mathcal{S}},t)$ and $\tilde{\Psi} g(\mathbf{x}_{\mathcal{S}},t)$: (a) Vessel phantom, (b) Palm18.} 
  \label{image:CoeffLog}
\end{figure}

Finally, we compare the efficacy of compression in data and image domain. To this end in Figure~\ref{image:CoeffLog}(a) we plot the decay of the log amplitudes of $\Psi g(t, \bx_{\mathcal{S}})$, $\tilde{\Psi}g(t, \bx_{\mathcal{S}})$ and $\Psi p_0^\textrm{TR}(g)$
noting that the cardinalities of image and data domain Curvelet transforms (before restriction) are similar but not the same as a result of the upscaling of the reconstruction grid. 
We see that the magnitudes of the image domain Curvelet coefficients, $\Psi p_0^\textrm{TR}(g)$, 
decay faster than the data domain coefficients $\Psi g(t, \bx_{\mathcal{S}})$ or $\tilde{\Psi}g(t, \bx_{\mathcal{S}})$ hinting at a higher compression rate and a more economic representation by Curvelets in image domain. We attribute this mainly to data, in contrast to initial pressure, not being compactly supported and hence more Curvelets needed to represent the hyperbolic branches. A tailored discretisation (progressively coarsening as $t$ increases) could somewhat counteract this issue but is out of scope of this paper. 

\subsubsection{Subsampling}
For such a compact phantom, uniform random subsampling of the flat sensor is not optimal. Instead we choose to concentrate the sampled points above the vessels. Practically, this is accomplished by sampling the centre of the sensor (the yellow window in Figure \ref{image:p0Vessel}) with probability 5 times higher, otherwise uniform, than the rest of the sensor.  
\begin{figure}[htbp!]
\vspace{-0.4cm} 
\setlength{\abovecaptionskip}{0.1cm} 
\setlength{\belowcaptionskip}{-1cm}   
\centering
  \subfloat[$b_0(t, \bx_{\mathcal{S}})$]{
  \includegraphics[width=0.48\linewidth,height=0.18\linewidth]{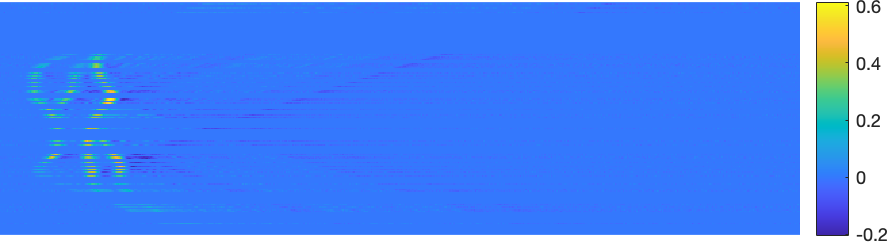}}
  \subfloat[$p_0^\textrm{TR}(b)$]{
  \includegraphics[width=0.5\linewidth,height=0.15\linewidth]{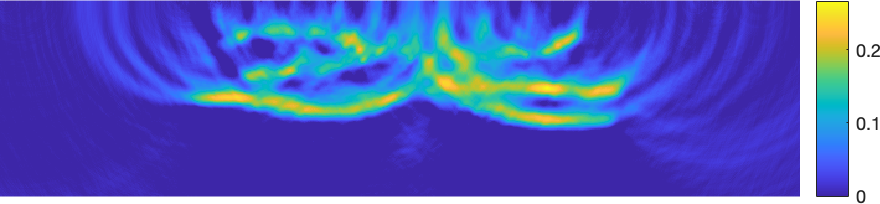}}
  \\
  \subfloat[$\hat{b}_0(t, \bx_{\mathcal{S}})$]{
  \includegraphics[width=0.48\linewidth,height=0.18\linewidth]{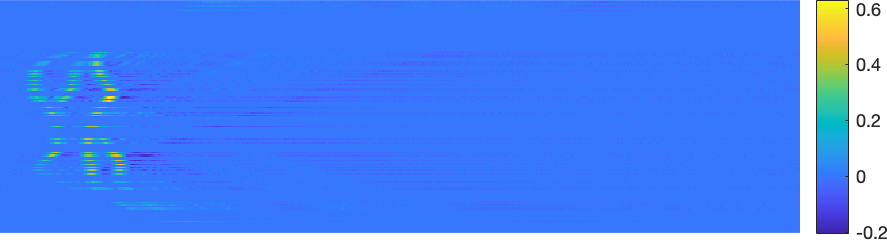}}
  \subfloat[$p_0^\textrm{TR}(\hat{b})$]{
  \includegraphics[width=0.5\linewidth,height=0.15\linewidth]{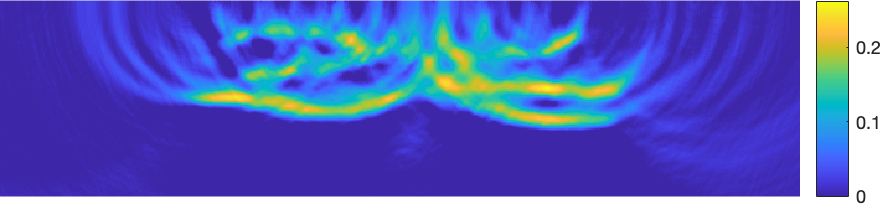}}
  \\
  \subfloat[ $\hat{b}_0(t, \bx_{\mathcal{S}}) - b_0(t, \bx_{\mathcal{S}})$]{
  \includegraphics[width=0.48\linewidth,height=0.18\linewidth]{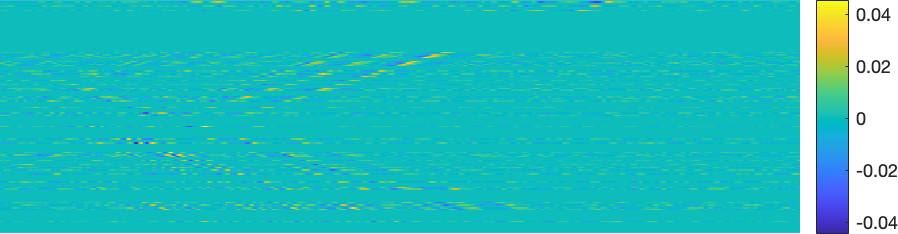}}
  \subfloat[$p_0^\textrm{TR}(\hat{b}) - p_0^\textrm{TR}(b)$]{
  \includegraphics[width=0.5\linewidth,height=0.15\linewidth]{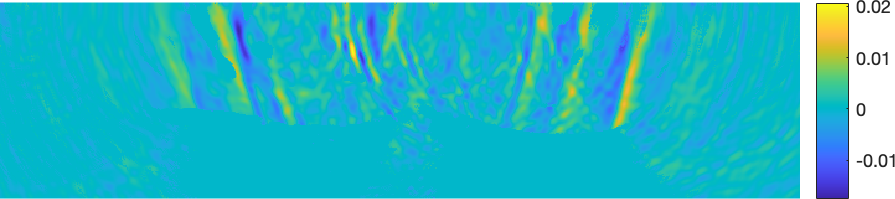}}
  \caption[VesselSSTR]{
  Counterpart of Figure \ref{image:VesselData} for subsampled data: (a) 25\% subsampled 0-filled data $b_0(t, \bx_{\mathcal{S}})$, (b) initial pressure via \ref{eq:TRequation} of $b$, (c) data projected on the range of $\tilde\Psi$, $\hat b_0(t, \bx_{\mathcal{S}}) = \tilde\Psi^\dagger \tilde\Psi b_0(t, \bx_{\mathcal{S}})$, (d) initial pressure via \ref{eq:TRequation} of $\hat b = \Phi\hat b_0(t, \bx_{\mathcal{S}})$, (e) data error $\hat{b}_0(t, \bx_{\mathcal{S}})$ - $b_0(t, \bx_{\mathcal{S}})$, (f) initial pressure error $p_0^\textrm{TR}(\hat{b}) - p_0^\textrm{TR}(b)$.}
  \label{image:VesselSS}
\end{figure}

\begin{figure}[htbp!]
\vspace{-0.3cm} 
\setlength{\abovecaptionskip}{0.1cm} 
\setlength{\belowcaptionskip}{-1cm}   
\centering
  \subfloat[$\Psi b_0(t, \bx_{\mathcal{S}})$]{
  \includegraphics[width=0.32\linewidth,height=0.2\linewidth]{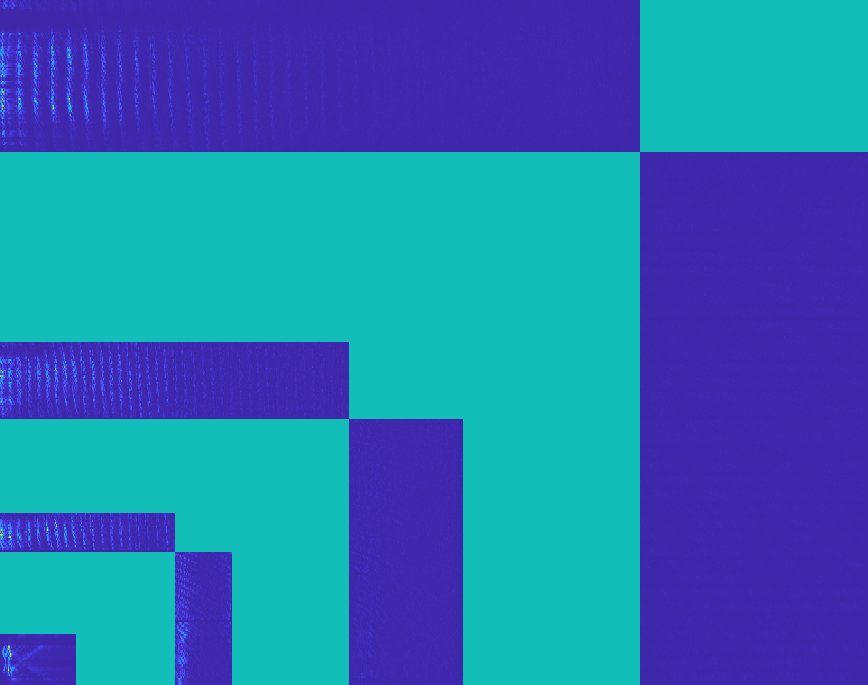}}
  \hspace{.2in}
  \subfloat[$\tilde{\Psi} b_0(t, \bx_{\mathcal{S}})$]{
  \includegraphics[width=0.32\linewidth,height=0.2\linewidth]{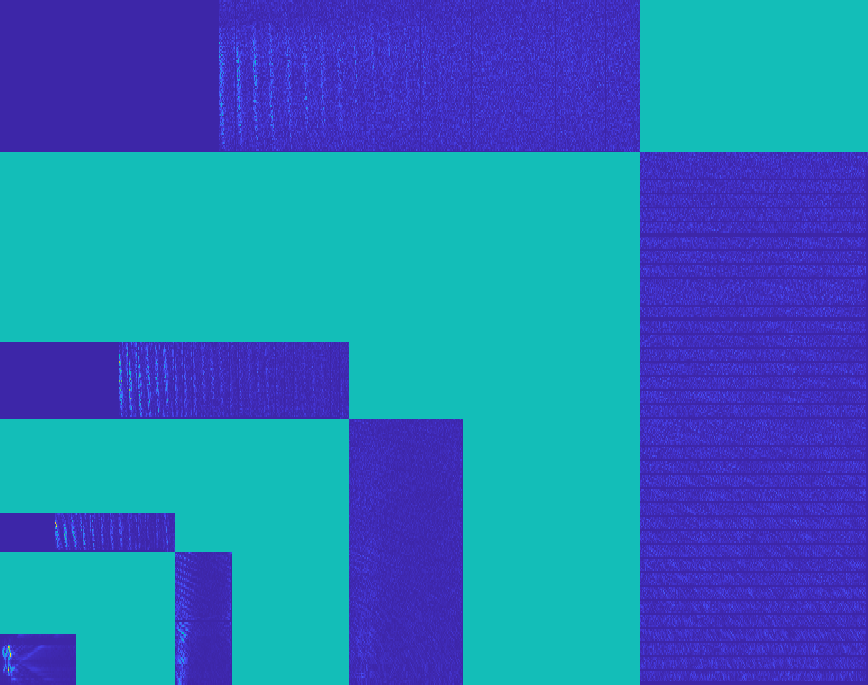}}
  \caption[VesselSSCurvelet]{Vessel phantom: Curvelet coefficients of subsampled, 0-filled data $b_0(t, \bx_{\mathcal{S}})$ (see Figure \ref{image:VesselCurvelet} for explanation of coefficient visualisation): (a) standard (4,152)-Curvelets with clearly visible out of range (non-physical) coefficients, (b) wedge restricted (4,128)-Curvelets.}
  \label{image:VesselSSCurvelet}
\end{figure}

Figure~\ref{image:VesselSS} is the counterpart of Figure~\ref{image:VesselData} for 25\% subsampled data $b$ visualised in Figure~\ref{image:VesselSS}(a) with the missing data replaced by 0 (to which we henceforth refer as $b_0(t, \bx_{\mathcal{S}})$, note that it holds $b = \Phi b_0$).
The time reversed solutions\footnote{We note that in \ref{eq:TRequation} we only impose the sampled sensor points i.e.~$b$, not the 0-filled data $b_0$, which leads to better results.} in (b) and (d) illustrate the subsampling artefacts in the linear reconstruction which are visible in the error plot (f) along the ``range discretisation'' artefacts already present in Figure~\ref{image:VesselData}(f). Plotting the Curvelet coefficients of $b_0$, Figure~\ref{image:VesselSSCurvelet}(a), reveals that the subsampling and 0-filling introduced out of range (non-physical) wavefronts perpendicular to the detector aligning with the jumps in $b_0(t, \bx_{\mathcal{S}})$ due to the all-0 data rows. These unwanted coefficients would derail the sparsity enhanced data reconstruction but are effectively removed by the wedge restricted transform, Figure~\ref{image:VesselSSCurvelet}(b), as expected without any perceptible detrimental effect on the data, Figure~\ref{image:VesselSS}(c,e). 

\begin{figure}[htbp!]
\vspace{-0.4cm} 
\setlength{\abovecaptionskip}{0.1cm} 
\setlength{\belowcaptionskip}{-1cm}   
  \hspace*{0.26in}
  \subfloat[$|\tilde{f}^\textrm{DR}|$]{
  \includegraphics[width=0.32\linewidth,height=0.2\linewidth]{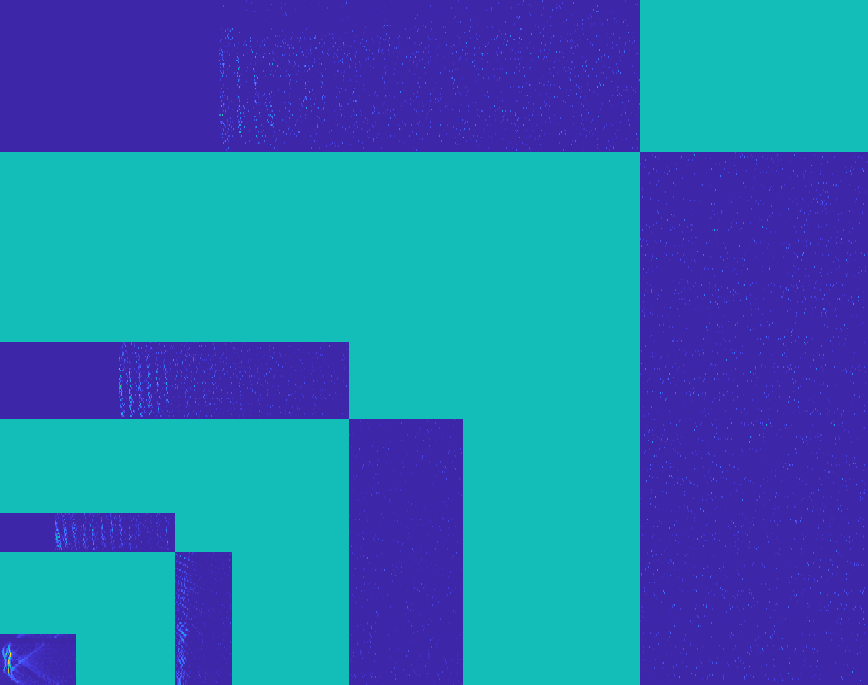}}
  \hspace*{0.3in}
  \subfloat[$\tilde{g}(t, \bx_{\mathcal{S}})$]{
  \includegraphics[width=0.48\linewidth,height=0.18\linewidth]{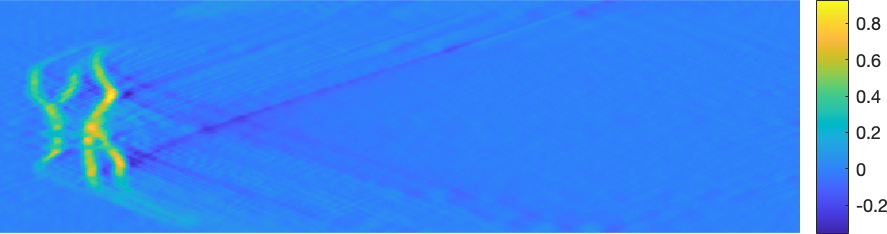}}\\
  \subfloat[$p_0^\textrm{TR}(\tilde{g})$]{
  \includegraphics[width=0.5\linewidth,height=0.15\linewidth]{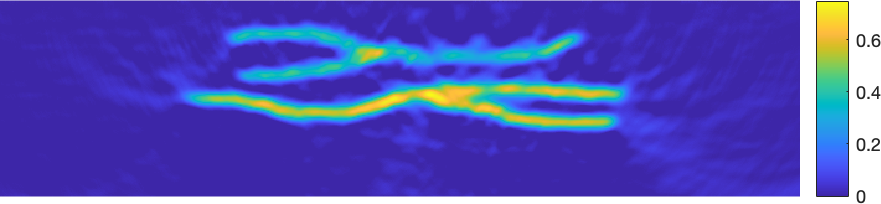}}
  \subfloat[$\tilde{g}(t, \bx_{\mathcal{S}})-g(t, \bx_{\mathcal{S}})$]{
  \includegraphics[width=0.48\linewidth,height=0.18\linewidth]{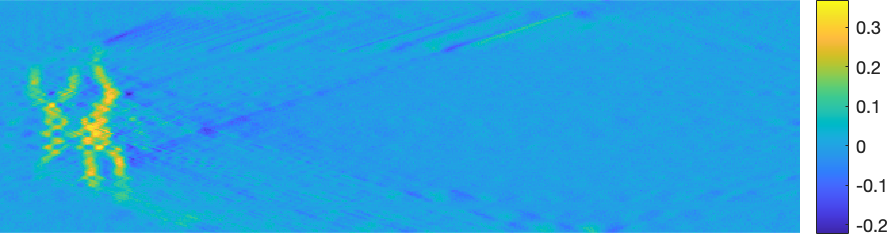}}
  \subfloat{
  \includegraphics[width=0.5\linewidth,height=0.15\linewidth]{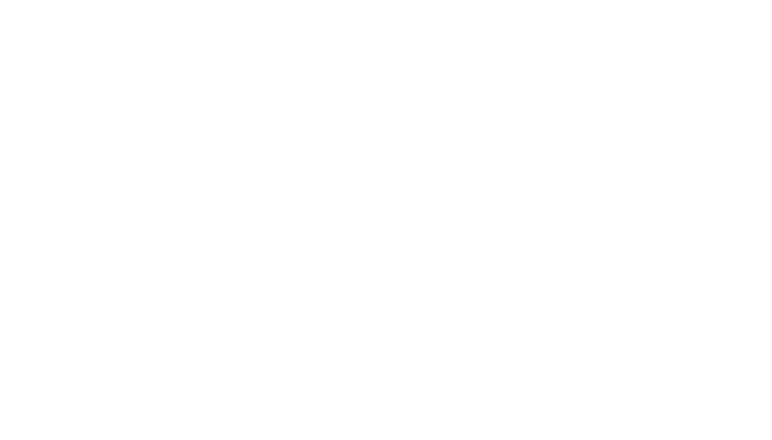}}
  \caption[VesselDR]{Vessel phantom: two step method. (a) Recovered wedge restricted Curvelet coefficients $\tilde{f}^\textrm{DR}$ (s.a.~description in Figure \ref{image:VesselCurvelet}), (b) the corresponding recovered full data $\tilde{g}(t, \bx_{\mathcal{S}})$, (c) initial pressure obtained via \ref{eq:TRequation} of $\tilde{g}(t, \bx_{\mathcal{S}})$, (d) data recovery error  $\tilde{g}(t, \bx_{\mathcal{S}}) - g(t, \bx_{\mathcal{S}})$.}
  \label{image:VesselDR}
\end{figure}

\subsubsection{Reconstruction}
\begin{figure}[htbp!]
\vspace{-0.3cm} 
\setlength{\abovecaptionskip}{0.1cm} 
\setlength{\belowcaptionskip}{-1cm}   
  \hspace{.26in}
  \subfloat[$|\Psi p_0^\textrm{R}(b)|$]{
  \includegraphics[width=0.32\linewidth,height=0.2\linewidth]{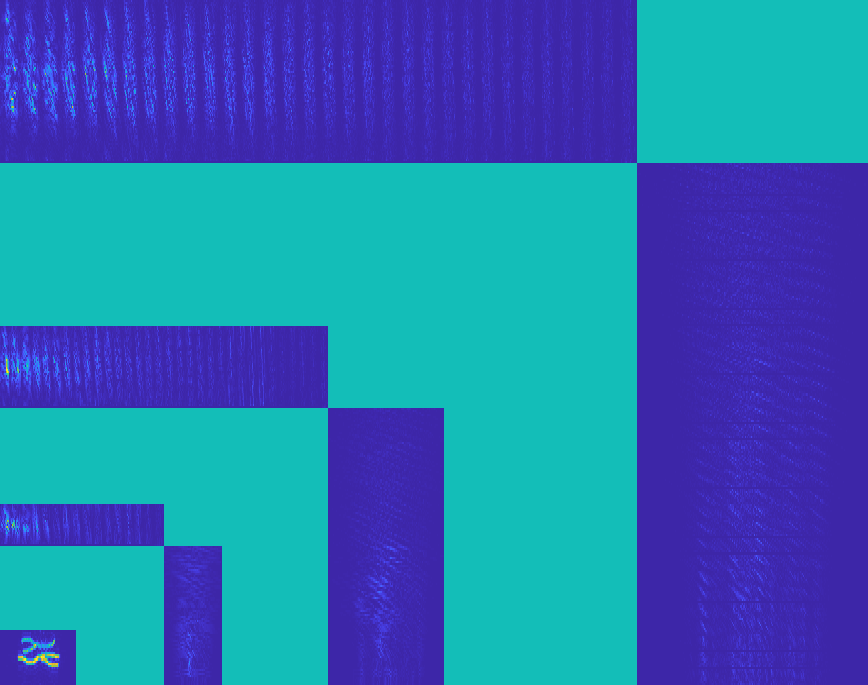}}
  \hspace{.52in}
  \subfloat[$|\Psi p_0^\textrm{R+}(b)|$]{
  \includegraphics[width=0.32\linewidth,height=0.2\linewidth]{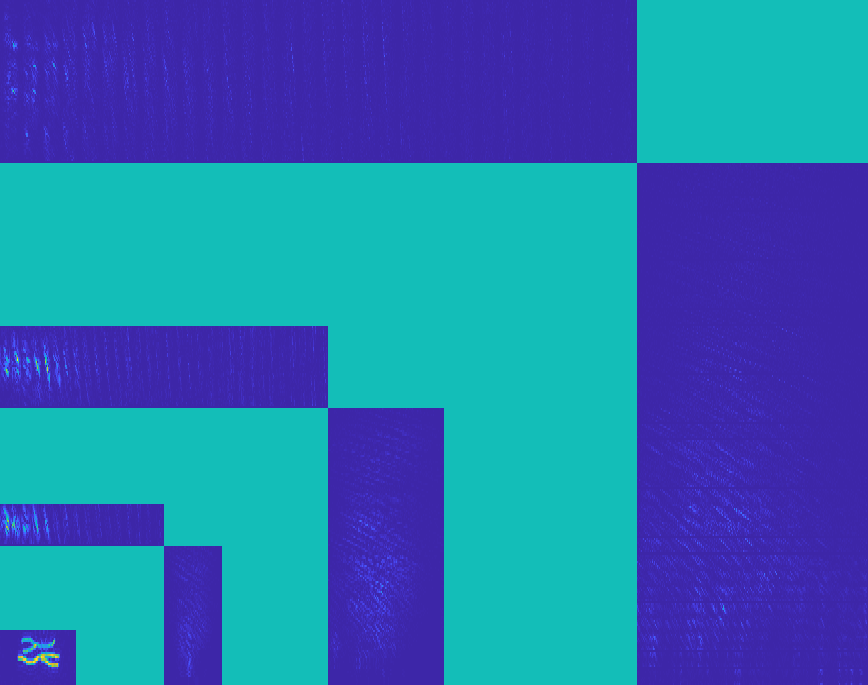}}
  \\
  \subfloat[$p_0^\textrm{R}(b)$]{
  \includegraphics[width=0.5\linewidth,height=0.15\linewidth]{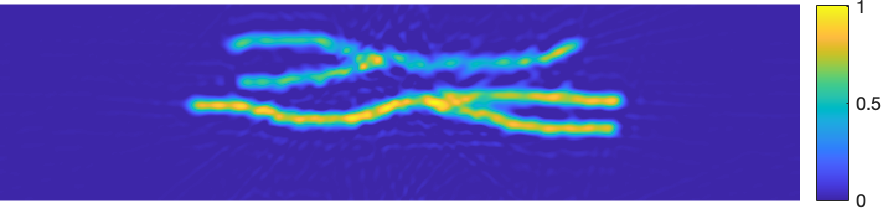}}
  \subfloat[$p_0^\textrm{R+}(b)$]{
  \includegraphics[width=0.5\linewidth,height=0.15\linewidth]{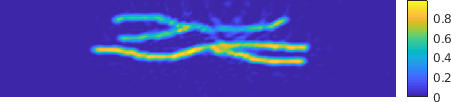}}
 \\
  \subfloat[${p_0}_{\lambda_1}^\textrm{TV+}(b)$]{
  \includegraphics[width=0.5\linewidth,height=0.15\linewidth]{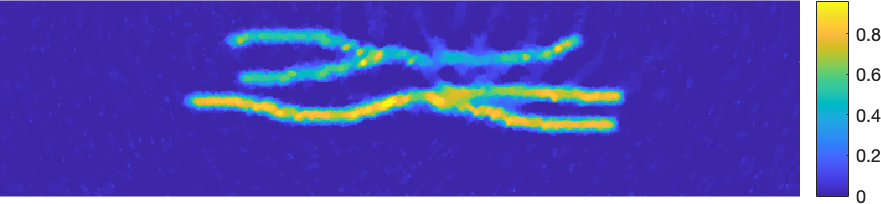}}
  \subfloat[${p_0}_{\lambda_2}^\textrm{TV+}(b)$]{
  \includegraphics[width=0.5\linewidth,height=0.15\linewidth]{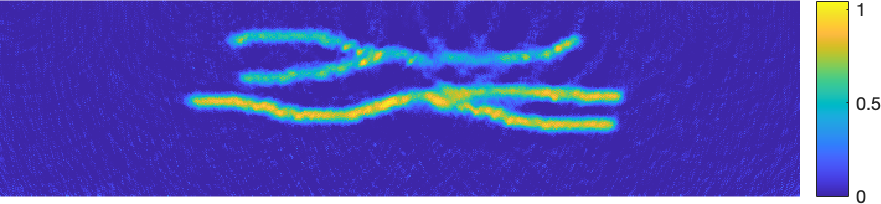}}
  \caption[Vesselp0Rp0TV]{Vessel phantom: one step methods \ref{p0R}, \ref{p0R+} (Curvelet sparsity), $p_0{\bf TV+}$ (total variation). (a,c)  $p_0^\textrm{R}(b)$ along its Curvelet coefficients (s.a.~description in Figure \ref{image:VesselCurvelet}); 
  (b,d) $p_0^\textrm{R+}(b)$ with non-negativity constraint along its Curvelet coefficients; 
  (e,f) $p_0^\textrm{TV+}(b)$ with ${\bf TV}$ regularisation and non-negativity constraint for regularisation parameter values $\lambda_1 = 5\cdot10^{-4}$ and $\lambda_2 = 10^{-5}$.}
  \label{image:Vesselp0Rp0TV}
\end{figure}

Finally, we compare performance of the two step and one step reconstruction methods on such 25\% subsampled data. 
The intermediate step of the two step procedure, the PAT data recovered with \textbf{R-SALSA} ($\tau=5\cdot 10^{-5}$, $\mu=1$, $C = 5$ and stopping tolerance $\eta =5\cdot10^{-4}$ or after 100 iterations) is visualised in Figure~\ref{image:VesselDR}:
(a) shows the recovered Curvelet coefficients, (b) the corresponding PAT data $\tilde{g}(t, \bx_{\mathcal{S}})$ and (d) the recovered data error $\tilde{g}(t, \bx_{\mathcal{S}}) - g(t, \bx_{\mathcal{S}})$.  
The initial pressure $p_0^\textrm{TR}(\tilde{g})$ obtained from the recovered PAT data $\tilde{g}(t, \bx_{\mathcal{S}})$ via time reversal is shown Figure~\ref{image:VesselDR}(c).
This can be compared against the one step procedure where the initial pressure $p_0$ is reconstructed directly from the subsampled data. Figure~\ref{image:Vesselp0Rp0TV} illustrates the results of the one step method in different scenarios: 
(a,c) $p_0^\textrm{R}(b)$ along its Curvelet coefficients reconstructed with \textbf{R-FISTA} ($\tau=10^{-3}$, $C = 5$ and stopping tolerance  $\eta=5\cdot10^{-4}$ or after 100 iterations);  
(b,d) $p_0^\textrm{R+}(b)$ along its Curvelet coefficients reconstructed with \textbf{R-ADMM} heeding the non-negativity constraint ($\tau=10^{-4}$, $\mu = 0.1$, $C = 5$ and stopping tolerance $\eta=5\cdot10^{-4}$ or after 100 iterations);
(e,f) for comparison purposes we also show result of reconstructions with non-negativity constraint total variation $\textbf{TV+}$, obtained with an ADMM method described in \cite{accelerate}, with two different values of the regularisation parameter $10^{-5}$ and $5\cdot 10^{-4}$.
The reconstructions obtained with two step method clearly exhibit more limited view artifacts. At least in part, we attribute this to the inefficiency of the data domain transform to represent exactly these peripheral wavefronts. The one step method produces overall superior results due to higher effectiveness of image domain Curvelet compression as well as iterated application of the forward and adjoint operators in the reconstruction process. The non-negativity constraint results in slightly smoother reconstructions. Despite having higher SNR, the $\textbf{TV+}$ reconstructions struggle to find a sweet spot between too noisy and overly smooth images while enforcing Curvelet sparsity results in a more continuous appearance of the reconstructed vessels. Quantitative comparison of the one and two step approaches including extra one step reconstructions \`a la \ref{p0R} with Haar and Daubechies-2 Wavelet bases is summarised in Table \ref{tab:quant:p0_basis}. The reconstructed images with Haar and Daubechies-2 Wavelet bases can be found in supplementary material. Despite Wavelets resulting in higher SSIM, the appearance of, in particular, the top vessel is less continuous in comparison with Curvelet reconstruction. Both Wavelet bases produce reconstructions with larger variation in contrast across the image hinting at isolated points with overestimated pressure, which are not apparent for reconstructions with Curvelet basis (which is consistent with its highest PSNR and MSE). 
To evaluate robustness to noise of our one and two step methods we reconstructed the same subsampled data but with higher noise levels. The results are summarised in Table \ref{tab:quant:noise} while images are again deferred to supplementary material. 
The imaging metrics are split. Uniformly lower MSE and higher PSNR are consistent with smoother images produced by the two step approach \ref{DR} while SSIM suggests that one step methods \ref{p0R}, \ref{p0R+} are better at preserving structure even for higher noise levels
which is corroborated by visual inspection of the reconstructed images (with exception of \ref{p0R+} for $\sigma= 0.08$).
\par
\begin{table*}[t]
\centering
\caption{Vessel Phantom: imaging metrics for $p_0$ reconstructed using different methods. The metrics are computed using the phantom image upscaled to the reconstruction grid via bilinear interpolation as ground truth.}
\begin{tabular}{c c|c|c|c|c|c|c|c|c|c|c}
    \toprule
    \midrule
    \multirow{2}[4]{*}{} & \multicolumn{2}{c}{Time reversal} & \multicolumn{1}{c}{\ref{DR} } & \multicolumn{8}{c}{Variational reconstruction \`a la \ref{p0R} with different bases: $p_0^{\textrm{crv}}(b)$, $p_0^{\textrm{crv+}}(b)$ correspond to \ref{p0R}, \ref{p0R+}.}\\
    \cmidrule(rl){2-3}\cmidrule(rl){4-4}\cmidrule(l){5-12}
    \bfseries & \bfseries $p_0^{\textrm{TR}}(g)$ & \bfseries $p_0^{\textrm{TR}}(b)$  & \bfseries $p_0^{\textrm{TR}}(\tilde{g})$ & \bfseries ${p_0}_{\lambda_1=10^{-5}}^\textrm{TV+}(b)$ & \bfseries ${p_0}_{\lambda_2=5\cdot10^{-4}}^\textrm{TV+}(b)$  & \bfseries $p_0^{\textrm{crv}}(b)$ & \bfseries $p_0^{\textrm{crv+}}(b)$ & \bfseries $p_0^{\textrm{db2}}(b)$ & \bfseries $p_0^{\textrm{db2+}}(b)$ & \bfseries $p_0^{\textrm{haar}}(b)$ & \bfseries $p_0^{\textrm{haar+}}(b)$\\ 
    \cmidrule(r){1-1}\cmidrule(rl){2-3}\cmidrule(rl){4-4}\cmidrule(l){5-12}
    \multicolumn{1}{l}{MSE}& 0.0153 & 0.0261 & 0.0107 & 0.004 & 0.0035 & 0.0034 & \textbf{0.0032} & 0.0042 & 0.0037 & 0.0058 & 0.0041\\
    \multicolumn{1}{l}{SSIM}& 0.6683 & 0.5532 & 0.6207 & 0.5329 & 0.7591 & 0.8079 & 0.791 & 0.8238 & 0.7312 & \textbf{0.8288} & 0.6862 \\
    \multicolumn{1}{l}{PSNR}& 18.1469 & 16.0234 & 18.7033 & 24.0295 & 24.5818 & 24.638 & \textbf{25.0053} & 23.7574 & 24.3298 & 22.3984 & 23.9174\\
    \multicolumn{1}{l}{SNR}& 18.8274 & 18.8169 & 22.8668 & \textbf{25.3583} & 25.2431 & 25.1421 & 25.2718 & 25.0987 & 25.2866 & 25.0487 & 25.3295  \\
    \midrule
    \bottomrule
\end{tabular}
\label{tab:quant:p0_basis}
\end{table*}

\begin{table}[t]
\centering
\caption{Vessel Phantom: imaging metrics for $p_0$ reconstructed with \ref{DR}, \ref{p0R}, \ref{p0R} for different noise levels. The metrics are computed using the phantom image upscaled to the reconstruction grid via bilinear interpolation as ground truth.}
\label{tab:quant:noise}
\begin{tabular}{c c|c|c|c|c|c}
    \toprule
    \midrule
    \multirow{2}[4]{*}{} & \multicolumn{3}{c}{$\sigma = 0.04$} & \multicolumn{3}{c}{$\sigma = 0.08$}\\ 
    \cmidrule(rl){2-4}\cmidrule(l){5-7}
    \bfseries &  \bfseries $p_0^{\textrm{TR}}(\tilde{g})$ & \bfseries $p_0^{\textrm{R}}(b)$ & \bfseries $p_0^{\textrm{R+}}(b)$ & \bfseries $p_0^{\textrm{TR}}(\tilde{g})$ & \bfseries $p_0^{\textrm{R}}(b)$ & \bfseries $p_0^{\textrm{R+}}(b)$ \\ 
    \cmidrule(r){1-1}\cmidrule(rl){2-4}\cmidrule(l){5-7}
    \multicolumn{1}{l}{MSE}& \textbf{0.0219} & 0.0324 & 0.0321 & \textbf{0.0226} & 0.0334 & 0.0328 \\
    \multicolumn{1}{l}{SSIM}& 0.5134 & 0.6117 & \textbf{0.6272} & 0.4163 & \textbf{0.4524} & 0.2421  \\
    \multicolumn{1}{l}{PSNR}& \textbf{16.5901} & 14.8992 & 14.9406 & \textbf{16.4616} & 14.7628 & 14.8392 \\
    \multicolumn{1}{l}{SNR}& 10.8592 & 13.1607 & \textbf{13.1989} & 4.9591 & 7.2005 & \textbf{7.3159} \\
    \midrule
    \bottomrule
\end{tabular}
\end{table}

%
%
\subsection{3D experimental data: palm18}\label{sec:PATRec:Palm1}
\begin{figure}[htbp!]
\vspace{-0.4cm} 
\setlength{\abovecaptionskip}{0.1cm} 
\setlength{\belowcaptionskip}{-1cm}   
  \centering
  \subfloat[left $g(t_{85},\bx_{\mathcal{S}})$, top right $g(t,\bx^{y_{59}}_S)$, bottom right $g(t,\mathbf{x}^{z_{59}}_S)$]{
  \begin{minipage}[b]{0.4\linewidth}
  \centering
  \includegraphics[width=0.7\textwidth,height=0.6\linewidth]{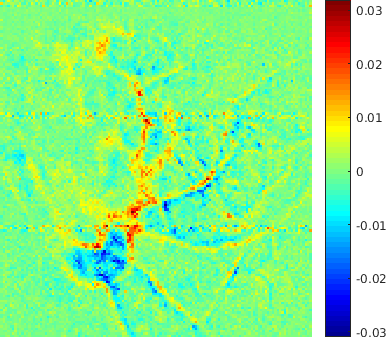}
  \centering
  \end{minipage}
  \begin{minipage}[b]{0.5\linewidth}
  \centering
  \includegraphics[width=1.05\textwidth,height=0.2\linewidth]{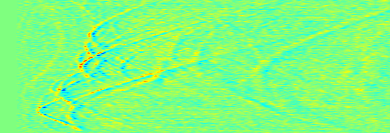}
  \vspace{0.0001in}
  \\
  \includegraphics[width=1.05\textwidth,height=0.2\linewidth]{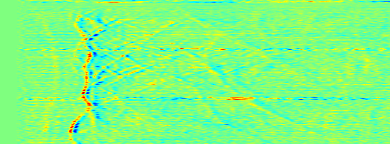}
  \end{minipage}}
  \\
  \centering
  \subfloat[left $\hat{g}(t_{85},\bx_{\mathcal{S}})$, top right $\hat{g}(t,\bx^{y_{59}}_S)$, bottom right $\hat{g}(t,\bx^{z_{59}}_S)$]{
  \begin{minipage}[b]{0.4\linewidth}
  \centering
  \includegraphics[width=0.7\textwidth,height=0.6\linewidth]{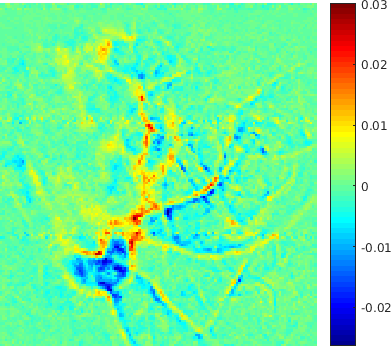}
  \end{minipage}
  \begin{minipage}[b]{0.5\linewidth}
  \centering
  \includegraphics[width=1.05\textwidth,height=0.2\linewidth]{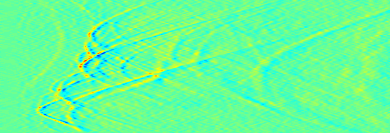}
  \vspace{0.0001in}
  \\
  \includegraphics[width=1.05\textwidth,height=0.2\linewidth]{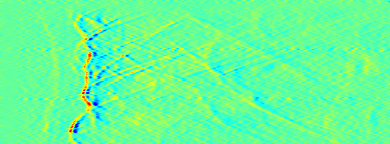}
  \end{minipage}}
  \caption[PalmData]{3D real data - palm18: cross-sections corresponding to time step $t = 85$, and scanning locations $y=59$, $z=59$ of (a) full data $g(t,\bx_{\mathcal{S}})$, (b) full data projected on the range of $\tilde\Psi$, $\hat g(t,\bx_{\mathcal{S}})=\tilde\Psi^\dagger \tilde\Psi g(t,\bx_{\mathcal{S}})$.
  }
  \label{image:PalmData}
\end{figure}

Finally, we compare the two and one step methods using Curvelets on a 3D data set of palm vasculature (palm18) acquired with a 16 beam Fabry-P\'erot scanner at 850nm excitation wavelength. The speed of sound and density are assumed homogeneous $c = 1570$m/s and 1000kg/m$^3$, respectively.
The full scan consists of a raster of $144\times133$ of equispaced locations with $h_\mathbf{x}=106\upmu\text{m}$, sampled every $h_t = 16.67$ns resulting in $n_t = 390$ time points. The palm18 data can be thought of as a time $\times$ sensor shaped volume with $390\times144\times133$ voxels with the sound speed per voxel $c_v = 0.2469$. Figure~\ref{image:PalmData}(a) shows the cross-sections of the palm18 data volume $g(\mathbf{x}_{\mathcal{S}},t)$ through time step $t=85$ and scanning locations $y=59$, $z=59$. 

\begin{figure}[htbp!]
\vspace{-0.4cm} 
\setlength{\abovecaptionskip}{0.1cm} 
\setlength{\belowcaptionskip}{-2cm}   
\centering
  \subfloat[$p_0^\textrm{TR}(g)$]{
  \includegraphics[width=0.31\linewidth,height=0.26\linewidth]{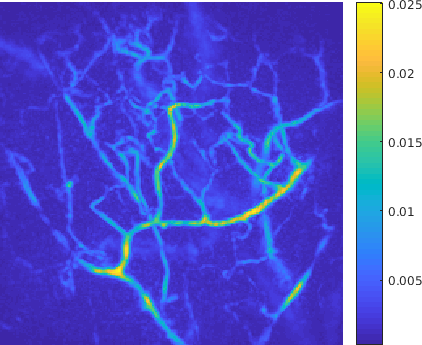}}
  \hfill
  \subfloat[$p_0^\textrm{TR}(\hat{g})$]{
  \includegraphics[width=0.31\linewidth,height=0.26\linewidth]{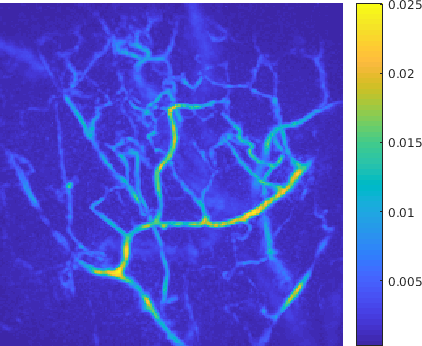}}
  \hfill
  \subfloat[$p_0^\textrm{TR}(\hat g) - p_0^\textrm{TR}(g)$]{
  \includegraphics[width=0.31\linewidth,height=0.27\linewidth]{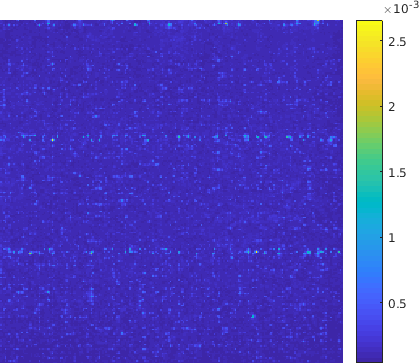}}
  \hfill\\
  \subfloat[$p_0^\textrm{TR}(b)$]{
  \includegraphics[width=0.31\linewidth,height=0.28\linewidth]{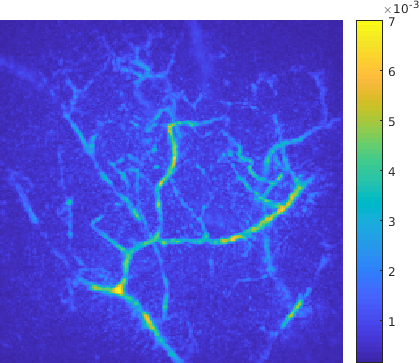}}
  \hfill
  \subfloat[$p_0^\textrm{TR}(\hat{b})$]{
  \includegraphics[width=0.31\linewidth,height=0.28\linewidth]{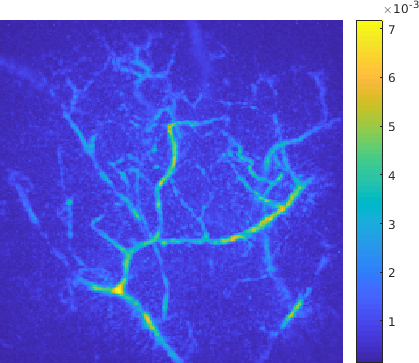}}
  \hfill
  \subfloat[$p_0^\textrm{TR}(\hat{b}) - p_0^\textrm{TR}(b)$]{
  \includegraphics[width=0.33\linewidth,height=0.28\linewidth]{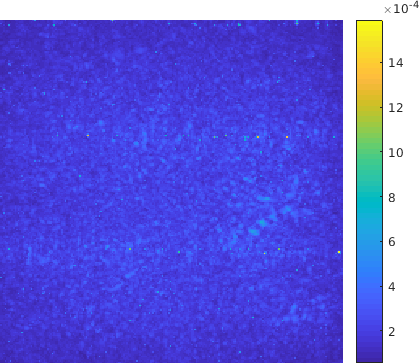}}
  \hfill 
  \caption[PalmTR]{3D real data - palm18: maximum intensity projection of initial pressure obtained via time reversal of (a) the full data $g(t, \bx_{\mathcal{S}})$, (b) the full data projected on the range of $\tilde\Psi$, $\hat{g}(t, \bx_{\mathcal{S}}) = \tilde\Psi^\dagger \tilde\Psi g(t, \bx_{\mathcal{S}})$ and (c) their difference $p_0^\textrm{TR}(\hat g) - p_0^\textrm{TR}(g)$; (d) the subsampled data $b$, (e) the subsampled data projected on the range of $\tilde\Psi$, $\hat{b} = \Phi\tilde\Psi^\dagger \tilde\Psi b_0(t, \bx_{\mathcal{S}})$ and (f) their difference $p_0^\textrm{TR}(\hat{b})-p_0^\textrm{TR}(b)$.}

  \label{image:PalmTR}
\end{figure}

Similarly as before, we upscale the reconstruction grid by the factor $\alpha = \sqrt[3]{n_t/n_{\bx_\perp}}$ to match the cardinality of the data as closely as possible. 
For the real data we calculate the image depth before upscaling as $n_{\bx_\perp} = n_t \cdot h_t \cdot c / h_\bx$ resulting in 
$\alpha = \sqrt[3]{h_\bx / (c \cdot h_t)} \approx 1.5941$ and the reconstruction grid $153\times229\times212$ with uniform spatial resolution of $66.5\upmu\text{m}$. Figure~\ref{image:PalmTR}(a) shows the the time reversal of the complete data on this grid which we use as a reference solution for the subsampled data reconstructions. 
\par

\dontshow{
\begin{figure}[htbp!]
\vspace{-0.6cm} 
\setlength{\abovecaptionskip}{0.1cm} 
\setlength{\belowcaptionskip}{-1cm}   
  \centering
  \includegraphics[width=0.55\linewidth,height=0.35\linewidth]{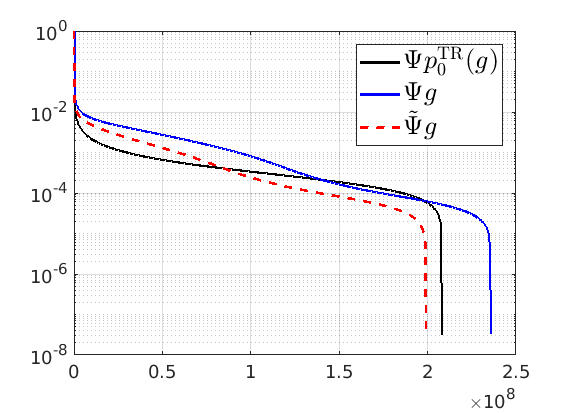}
  \caption[PalmLog]{3D real data - palm18: the decay of log amplitudes of image and data Curvelet coefficients: $\Psi p_0^\textrm{TR}(g)$, $\Psi g(t, \bx_{\mathcal{S}})$ and $\tilde{\Psi} g(t, \bx_{\mathcal{S}})$.}
  \label{image:PalmLog}
\end{figure}
}
We construct the image Curvelets $\Psi$ and the data wedge restricted Curvelets $\tilde\Psi$ (after restriction) to both have 4 scales and 128 angles (at the 2nd coarsest level in each plane) and conduct analogous experiments as in 2D. Figure~\ref{image:PalmData}(b) shows the data projected on the range of $\tilde\Psi$, $\hat g(t,\bx_{\mathcal{S}}) = \tilde\Psi^\dagger \tilde\Psi g(t,\bx_{\mathcal{S}})$. We note that the projection effectively removes the faulty sensors appearing as horizontal lines in the bottom plot on the right.

Figure~\ref{image:PalmTR}(b) shows $p_0^\textrm{TR}(\hat{g})$ obtained via the time reversal of the projected data $\hat g(t,\bx_{\mathcal{S}})$ and (c) the difference in image domain $p_0^\textrm{TR}(\hat{g}) - p_0^\textrm{TR}(g)$.  
The log magnitude plot of Curvelet coefficients in Figure~\ref{image:CoeffLog}(b) reveals that while the magnitudes of the image domain Curvelet coefficients $\Psi p_0^\textrm{TR}(g)$ decay the fastest, the magnitudes of wedge restricted Curvelet coefficients $\tilde{\Psi} g(t,\bx_{\mathcal{S}})$ still decay at a faster rate than $\Psi g(t,\bx_{\mathcal{S}})$, which was not the case in 2D.\par
\begin{figure}[htbp!]
\vspace{-0.2cm} 
\setlength{\abovecaptionskip}{0.1cm} 
\setlength{\belowcaptionskip}{-1cm}   
  \centering
  \subfloat[left $b_0(t_{85},\bx_{\mathcal{S}})$, top right $b_0(t,\bx_{\mathcal{S}}^{y_{59}})$, bottom right $b_0(t,\bx_{\mathcal{S}}^{z_{59}})$]{
  \begin{minipage}[b]{0.4\linewidth}
  \centering
  \includegraphics[width=0.7\textwidth,height=0.6\linewidth]{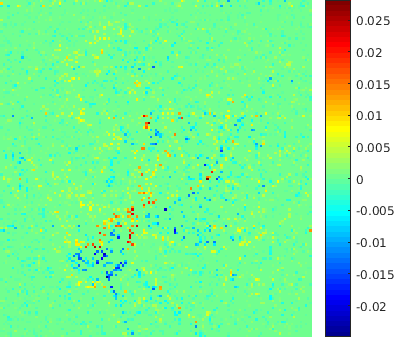}
  \centering
  \end{minipage}
  \begin{minipage}[b]{0.5\linewidth}
  \centering
  \includegraphics[width=1.05\textwidth,height=0.2\linewidth]{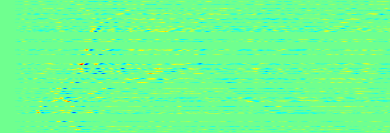}
  \vspace{0.0001in}
  \\
  \centering
  \includegraphics[width=1.05\textwidth,height=0.2\linewidth]{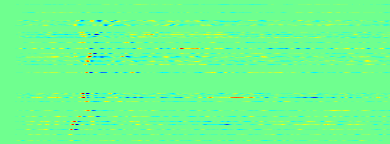}
  \end{minipage}}
  \\
  \centering
  \subfloat[left $\tilde{g}(t_{85},\bx_{\mathcal{S}})$, top right $\tilde{g}(t,\bx_\mathcal{S}^{y_{59}})$, bottom right $\tilde{g}(t,\bx_{\mathcal{S}}^{z_{59}})$]{
  \begin{minipage}[b]{0.4\linewidth}
  \centering
  \includegraphics[width=0.7\textwidth,height=0.6\linewidth]{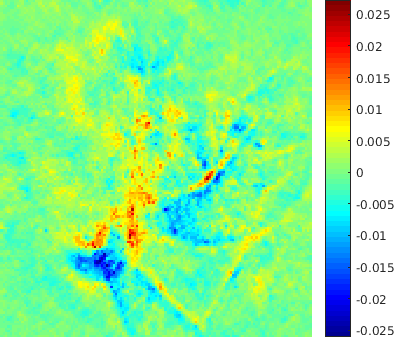}
  \centering
  \end{minipage}
  \begin{minipage}[b]{0.5\linewidth}
  \centering
  \includegraphics[width=1.05\textwidth,height=0.2\linewidth]{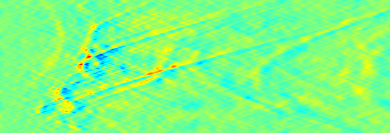}
  \vspace{0.0001in}
  \\
  \includegraphics[width=1.05\textwidth,height=0.2\linewidth]{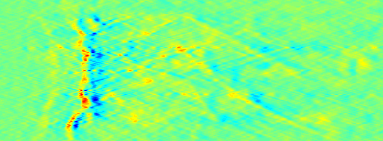}
  \end{minipage}}
  \\
  \centering
  \subfloat[$\tilde{g}(t,\bx_{\mathcal{S}}) - \hat{g}(t,\bx_{\mathcal{S}})$ at $t_{85}$, $\bx_{\mathcal{S}}^{y_{59}}$ and $\bx_{\mathcal{S}}^{z_{59}}$]{
  \begin{minipage}[b]{0.4\linewidth}
  \centering
  \includegraphics[width=0.7\textwidth,height=0.6\linewidth]{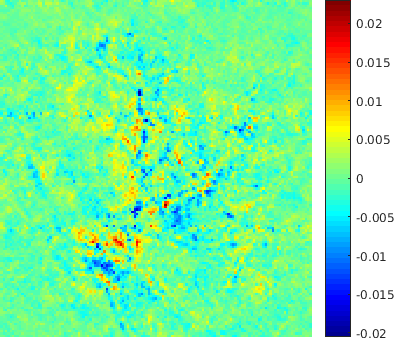}
  \centering
  \end{minipage}
  \begin{minipage}[b]{0.5\linewidth}
  \centering
  \includegraphics[width=1.05\textwidth,height=0.2\linewidth]{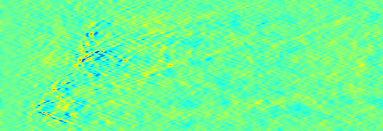}
  \vspace{0.0001in}
  \\
  \includegraphics[width=1.05\textwidth,height=0.2\linewidth]{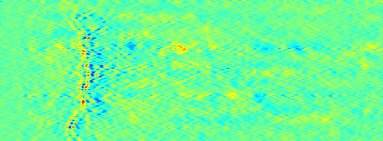}
  \end{minipage}}
  \caption[PalmDR]{3D real data - palm18: cross-sections corresponding to time step $t = 85$, and scanning locations $y=59$, $z=59$ of (a) 25\% uniformly randomly subsampled, 0-filled data $b_0(t,\bx_{\mathcal{S}})$, (b) full data $\tilde{g}(t,\bx_{\mathcal{S}})$ recovered with \textbf{R-SALSA}, (c) $\tilde\Psi$ range restricted data recovery error $\tilde{g}(t,\bx_{\mathcal{S}})-\hat{g}(t,\bx_{\mathcal{S}})$.}
  \label{image:PalmDR}
\end{figure}

Next, we uniformly randomly subsample the data down to 25\%  resulting in, after 0-filling, $b_0(t,\bx_{\mathcal{S}})$ which cross-sections at time step $t=85$ and scanning locations $y=59$, $z=59$ are shown in Figure~\ref{image:PalmDR}(a). 
Figure~\ref{image:PalmTR}(f) 
illustrates the noise-like difference between the initial pressure obtained via time reversal of the subsampled data $b$ shown in Figure~\ref{image:PalmTR}(d) and the subsampled, 0-filled data projected on the range of $\tilde\Psi$, $\hat b = \Phi \tilde\Psi^\dagger\tilde\Psi b_0(t,\bx_{\mathcal{S}})$ shown in Figure~\ref{image:PalmTR}(e).

\begin{figure}[htbp!]
\vspace{-0.6cm} 
\setlength{\abovecaptionskip}{0.1cm} 
\setlength{\belowcaptionskip}{-1cm}   
\centering
  \subfloat[$p_0^\textrm{TR}(\tilde{g})$]{
  \includegraphics[width=0.3\linewidth,height=0.26\linewidth]{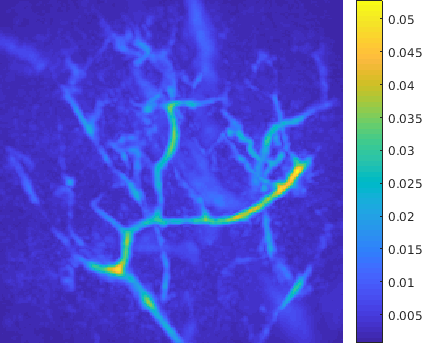}}
  \hfill
  \subfloat[$p_0^\textrm{R}(b)$]{
  \includegraphics[width=0.3\linewidth,height=0.26\linewidth]{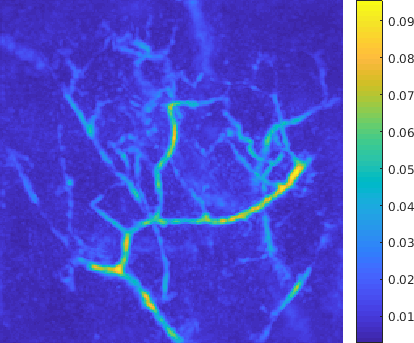}}
  \hfill
  \subfloat[$p_0^\textrm{R+}(b)$]{
  \includegraphics[width=0.3\linewidth,height=0.26\linewidth]{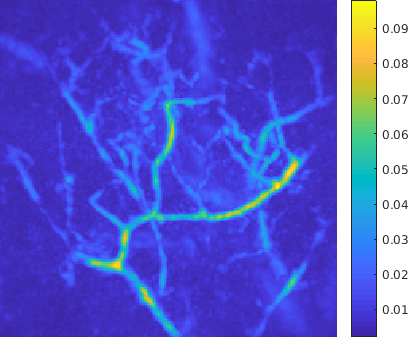}}
  \\
  \centering
  \subfloat[${p_0}_{\lambda_1}^\textrm{TV+}(b)$]{
  \includegraphics[width=0.31\linewidth,height=0.28\linewidth]{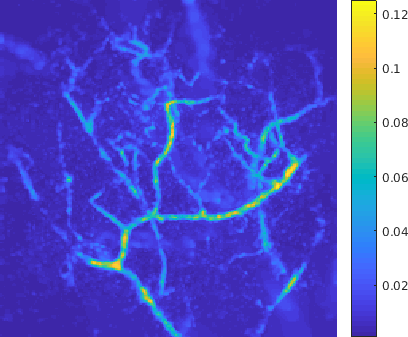}}
  \hfill
  \subfloat[${p_0}_{\lambda_2}^\textrm{TV+}(b)$]{
  \includegraphics[width=0.31\linewidth,height=0.28\linewidth]{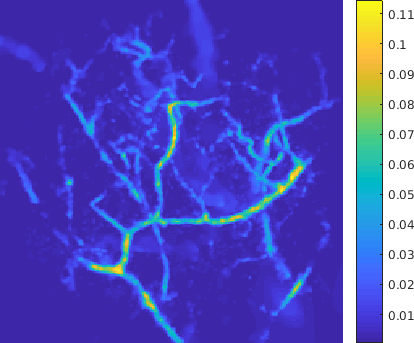}}
  \hspace*{1.2in}
  \caption[Palmp0R]{3D real data - palm18: maximum intensity projection of initial pressure (a) obtained via \ref{eq:TRequation} from recovered PAT data $\tilde{g}(\mathbf{x}_{\mathcal{S}},t)$, (b) $p_0^\textrm{R}(b)$ via one step method \ref{p0R}, (c) $p_0^\textrm{R+}(b)$ via one step method \ref{p0R+}, (d,e) $p_0^\textrm{TV+}(b)$ via one step method \textbf{TV+} for regularisation parameter values $\lambda_1 = 2.5\cdot10^{-4}$ and $\lambda_2 = 5\cdot10^{-4}$.}
  \label{image:PalmDRp0Rp0TV}
\end{figure}

The 3D data $\tilde{g}(t,\bx_{\mathcal{S}})$ recovered with \textbf{R-SALSA} ($\tau=1\cdot 10^{-5}$, $\mu=1$, $C = 5$ and stopping tolerance $\eta =5\cdot10^{-4}$ or after 50 iterations) in the first step of the two step method is shown in Figure~\ref{image:PalmDR}(b) and its range restricted error in (c). Time reversal of $\tilde{g}(t,\bx_{\mathcal{S}})$ results in initial pressure $p_0^\textrm{TR}(\tilde{g})$ in Figure~\ref{image:PalmDRp0Rp0TV}(a). 
The one step $p_0^\textrm{R}(b)$ reconstruction with \textbf{R-FISTA} ($\tau=2.5\cdot 10^{-6}$, $C = 5$ and stopping tolerance  $\eta=5\cdot10^{-4}$ or after 50 iterations) is illustrated in Figure~\ref{image:PalmDRp0Rp0TV}(b) and $p_0^\textrm{R+}(b)$ its counterpart with non-negativity constraint reconstructed with \textbf{R-ADMM} ($\tau=2.5\cdot 10^{-6}$, $\mu = 0.1$, $C = 5$ and stopping tolerance $\eta=5\cdot10^{-4}$ or after 50 iterations) in Figure~\ref{image:PalmDRp0Rp0TV}(c). Again, for comparison in Figure~\ref{image:PalmDRp0Rp0TV}(d,e) we plot one step $p_0$ reconstructions with non-negativity constraint and total variation regularisation $\textbf{TV+}$ (see \cite{accelerate}) for a range of regularization parameters.\par

More aggressive subsampling with 12.5\% data results in a subtle deprecation of the quality of reconstructed images for all methods: higher background noise, smaller/thinner vessels brake up, more blur (see supplementary material).

%
%
\section{Conclusion}\label{sec:Conclusion}
Summarising, we derived a one-to-one map between the wavefront directions in ambient (and hence image) space and data space which suggested a near equivalence of initial pressure and PAT data volume reconstructions assuming sparsity in Curvelet frame. We introduced a novel wedge restriction to the Curvelet transform which allowed us to define a tight Curvelet frame on the range of PAT forward operator which is crucial to successful full data reconstruction from subsampled data. Intrigued by this near equivalence we compared one and two step approaches to photoacoustic image reconstruction from subsampled data utilising sparsity in tailored Curvelet frames. The major benefit of any two step approach is that it decouples the nonlinear iterative compressed sensing recovery and the linear acoustic propagation problems, thus the expensive acoustic forward and adjoint operators are not iterated with. A further important advantage w.r.t.~an earlier two step method \cite{Acoustic} is exploitation of the relationship between the time steps (as it is the case for one step methods) by treating the entire PAT data as a volume which is then transformed with the wedge restricted Curvelet frame. 

The main drawbacks of the data reconstruction are, that the PAT geometry results in a larger data volume than the original PAT image which is further exacerbated by the commonly employed time oversampling of the PAT data. In effect, we have to deal with a larger volume and consequently more coefficients in compressed sensing recovery. Furthermore flat sensor PAT data is effectively unbounded (only restricted by sensor size) even for compactly supported initial pressures leading to less sparse/compressed representations. This is reflected in our experiments indicating that PAT data is less compressible in the wedge restricted Curvelet frame than PAT image in the standard Curvelet frame which for the same number of measurements yields lower quality data reconstructions than those obtained with one step methods.\par 
Comparing one step methods \ref{p0R}, \ref{p0R+} to total variation regularisation \textbf{TV+} or \`a la \ref{p0R} reconstructions with Wavelet bases, Curvelet sparsity constraint results in smoother looking vessel images. The reason is that vasculature images commonly contain wide range of scales (vessel diameters) to which 
multiscale frames can easily adapt. While Wavelet bases share this property, they cannot, in contrast to Curvelets, represent curve like singularities which are common feature of vascular images.
On the other hand, no single regularisation parameter value for total variation can effectively denoise larger features without removing/blurring finer ones. The main drawback of the one step methods is their computational cost due to iterative evaluation of the expensive forward and adjoint operators, in particular when using the pseudospectral method which cannot take advantage of sparse sampling. Other forward solvers based e.g.~on ray tracing could potentially be used to alleviate some of the costs \cite{rullan2018hamilton}, \cite{rullan2020photoacoustic}.\par
Finally, we would like to note that Curvelets provide a Wavelet-Vagulette 
biorthogonal decomposition for the Radon transform and similar result was shown for PAT in \cite{frikel2020sparse}.
The wavefront mapping in Section \ref{sec:Curvelet:WaveFrontMapping} provides a relationship between the image and data domain wavefronts and hence the corresponding Curvelets. However, the proposed reconstruction methods are not based on Wavelet-Vagulette decomposition which is subject of future work.

\section{Acknowledgements}\label{sec:Acknow}
The authors acknowledge financial support by EPSRC grant EP/K009745/1 and ERC grant 74119.   


%





\ifCLASSOPTIONcaptionsoff
  \newpage
\fi



%
%
%
\bibliographystyle{IEEEtran}
\bibliography{bibliography.bib}

\begin{thebibliography}{10}
\providecommand{\url}[1]{#1}
\csname url@samestyle\endcsname
\providecommand{\newblock}{\relax}
\providecommand{\bibinfo}[2]{#2}
\providecommand{\BIBentrySTDinterwordspacing}{\spaceskip=0pt\relax}
\providecommand{\BIBentryALTinterwordstretchfactor}{4}
\providecommand{\BIBentryALTinterwordspacing}{\spaceskip=\fontdimen2\font plus
\BIBentryALTinterwordstretchfactor\fontdimen3\font minus
  \fontdimen4\font\relax}
\providecommand{\BIBforeignlanguage}[2]{{%
\expandafter\ifx\csname l@#1\endcsname\relax
\typeout{** WARNING: IEEEtran.bst: No hyphenation pattern has been}%
\typeout{** loaded for the language `#1'. Using the pattern for}%
\typeout{** the default language instead.}%
\else
\language=\csname l@#1\endcsname
\fi
#2}}
\providecommand{\BIBdecl}{\relax}
\BIBdecl

\bibitem{NearOptimal}
E.~J. Cand{\`e}s and T.~Tao, ``Near-optimal signal recovery from random
  projections: Universal encoding strategies?'' \emph{IEEE Transactions on
  Information Theory}, vol.~52, no.~12, pp. 5406--5425, 2006.

\bibitem{Robust}
E.~J. Cand{\`e}s, J.~Romberg, and T.~Tao, ``Robust uncertainty principles:
  Exact signal reconstruction from highly incomplete frequency information,''
  \emph{IEEE Transactions on Information Theory}, vol.~52, no.~2, pp. 489--509,
  2006.

\bibitem{stableRecovery}
E.~J. Cand{\`e}s, J.~K. Romberg, and T.~Tao, ``Stable signal recovery from
  incomplete and inaccurate measurements,'' \emph{Communications on Pure and
  Applied Mathematics}, vol.~59, no.~8, pp. 1207--1223, 2006.

\bibitem{donohoCS}
D.~L. Donoho, ``Compressed sensing,'' \emph{IEEE Transactions on Information
  Theory}, vol.~52, no.~4, pp. 1289--1306, 2006.

\bibitem{NC}
E.~J. Cand{\`e}s and D.~L. Donoho, ``New tight frames of curvelets and optimal
  representations of objects with piecewise ${C}^2$ singularities,''
  \emph{Communications on Pure and Applied Mathematics}, vol.~57, no.~2, pp.
  219--266, 2004.

\bibitem{dataMotivation}
E.~J. Cand{\`e}s and L.~Demanet, ``The curvelet representation of wave
  propagators is optimally sparse,'' \emph{Communications on Pure and Applied
  Mathematics}, vol.~58, no.~11, pp. 1472--1528, 2005.

\bibitem{PATsys}
E.~Zhang, J.~Laufer, and P.~Beard, ``Backward-mode multiwavelength
  photoacoustic scanner using a planar {Fabry-Perot} polymer film ultrasound
  sensor for high-resolution three-dimensional imaging of biological tissues,''
  \emph{Applied Optics}, vol.~47, no.~4, pp. 561--577, 2008.

\bibitem{SubsampledFP17}
N.~Huynh, F.~Lucka, E.~Zhang, M.~Betcke, S.~Arridge, P.~Beard, and B.~Cox,
  ``Sub-sampled {Fabry Perot} photoacoustic scanner for fast {3D} imaging,'' in
  \emph{Photons Plus Ultrasound: Imaging and Sensing 2017}, vol. 10064, 2017,
  p. 100641Y.

\bibitem{PatternPAT14}
N.~Huynh, E.~Zhang, M.~Betcke, S.~Arridge, P.~Beard, and B.~Cox, ``Patterned
  interrogation scheme for compressed sensing photoacoustic imaging using a
  {Fabry Perot} planar sensor,'' in \emph{Photons Plus Ultrasound: Imaging and
  Sensing 2014}, vol. 8943, 2014, p. 894327.

\bibitem{RTUSField15}
N.~Huynh, E.~Zhang, M.~Betcke, S.~R. Arridge, P.~Beard, and B.~Cox, ``A
  real-time ultrasonic field mapping system using a {Fabry Perot} single pixel
  camera for {3D} photoacoustic imaging,'' in \emph{Photons Plus Ultrasound:
  Imaging and Sensing 2015}, vol. 9323, 2015, p. 93231O.

\bibitem{SPOC}
N.~Huynh, E.~Zhang, M.~Betcke, S.~Arridge, P.~Beard, and B.~Cox, ``Single-pixel
  optical camera for video rate ultrasonic imaging,'' \emph{Optica}, vol.~3,
  no.~1, pp. 26--29, Jan 2016.

\bibitem{Acoustic}
M.~M. Betcke, B.~T. Cox, N.~Huynh, E.~Z. Zhang, P.~C. Beard, and S.~R. Arridge,
  ``Acoustic wave field reconstruction from compressed measurements with
  application in photoacoustic tomography,'' \emph{IEEE Transactions on
  Computational Imaging}, vol.~3, no.~4, pp. 710--721, 2017.

\bibitem{ACSPAT}
J.~Provost and F.~Lesage, ``The application of compressed sensing for
  photo-acoustic tomography,'' \emph{IEEE Transactions on Medical Imaging},
  vol.~28, no.~4, pp. 585--594, 2009.

\bibitem{CSPATinvivo}
Z.~Guo, C.~Li, L.~Song, and L.~V. Wang, ``Compressed sensing in photoacoustic
  tomography in vivo,'' \emph{Journal of Biomedical Optics}, vol.~15, no.~2, p.
  021311, 2010.

\bibitem{FullWave}
C.~Huang, K.~Wang, L.~Nie, L.~V. Wang, and M.~A. Anastasio, ``Full-wave
  iterative image reconstruction in photoacoustic tomography with acoustically
  inhomogeneous media,'' \emph{IEEE Transactions on Medical Imaging}, vol.~32,
  no.~6, pp. 1097--1110, 2013.

\bibitem{FBEMadjoint}
Z.~Belhachmi, T.~Glatz, and O.~Scherzer, ``A direct method for photoacoustic
  tomography with inhomogeneous sound speed,'' \emph{Inverse Problems},
  vol.~32, no.~4, p. 045005, 2016.

\bibitem{PATadjoint}
S.~R. Arridge, M.~M. Betcke, B.~T. Cox, F.~Lucka, and B.~E. Treeby, ``On the
  adjoint operator in photoacoustic tomography,'' \emph{Inverse Problems},
  vol.~32, no.~11, p. 115012, 2016.

\bibitem{accelerate}
S.~Arridge, P.~Beard, M.~Betcke, B.~Cox, N.~Huynh, F.~Lucka, O.~Ogunlade, and
  E.~Zhang, ``Accelerated high-resolution photoacoustic tomography via
  compressed sensing,'' \emph{Physics in Medicine \& Biology}, vol.~61, no.~24,
  p. 8908, 2016.

\bibitem{javaherian2016multi}
A.~\text{}Javaherian and S.~Holman, ``A multi-grid iterative method for
  photoacoustic tomography,'' \emph{IEEE transactions on medical imaging},
  vol.~36, no.~3, pp. 696--706, 2016.

\bibitem{haltmeier2017analysis}
M.~Haltmeier and L.~V. Nguyen, ``Analysis of iterative methods in photoacoustic
  tomography with variable sound speed,'' \emph{SIAM Journal on Imaging
  Sciences}, vol.~10, no.~2, pp. 751--781, 2017.

\bibitem{CSSPAT}
M.~Haltmeier, T.~Berer, S.~Moon, and P.~Burgholzer, ``Compressed sensing and
  sparsity in photoacoustic tomography,'' \emph{Journal of Optics}, vol.~18,
  no.~11, p. 114004, 2016.

\bibitem{xu2005universal}
M.~Xu and L.~V. Wang, ``Universal back-projection algorithm for photoacoustic
  computed tomography,'' \emph{Physical Review E}, vol.~71, no.~1, p. 016706,
  2005.

\bibitem{wang2011photoacoustic}
K.~Wang and M.~A. Anastasio, ``Photoacoustic and thermoacoustic tomography:
  image formation principles,'' in \emph{Handbook of Mathematical Methods in
  Imaging}.\hskip 1em plus 0.5em minus 0.4em\relax Springer, 2011, pp.
  781--815.

\bibitem{PATbackward}
D.~Finch and S.~K. Patch, ``Determining a function from its mean values over a
  family of spheres,'' \emph{SIAM Journal on Mathematical Analysis}, vol.~35,
  no.~5, pp. 1213--1240, 2004.

\bibitem{StefanovUhlman}
P.~Stefanov and G.~Uhlmann, ``Thermoacoustic tomography with variable sound
  speed,'' \emph{Inverse Problems}, vol.~25, no.~7, p. 075011, 2009.

\bibitem{kWave}
B.~E. Treeby and B.~T. Cox, ``k-wave: Matlab toolbox for the simulation and
  reconstruction of photoacoustic wave fields,'' \emph{Journal of Biomedical
  Optics}, vol.~15, no.~2, p. 021314, 2010.

\bibitem{cox2005fast}
B.~Cox and P.~Beard, ``Fast calculation of pulsed photoacoustic fields in
  fluids using k-space methods,'' \emph{The Journal of the Acoustical Society
  of America}, vol. 117, no.~6, pp. 3616--3627, 2005.

\bibitem{TBPO}
K.~P. K{\"o}stli, M.~Frenz, H.~Bebie, and H.~P. Weber, ``Temporal backward
  projection of optoacoustic pressure transients using {F}ourier transform
  methods,'' \emph{Physics in Medicine \& Biology}, vol.~46, no.~7, p. 1863,
  2001.

\bibitem{EFRFTT}
Y.~Xu, D.~Feng, and L.~V. Wang, ``Exact frequency-domain reconstruction for
  thermoacoustic tomography. {I. Planar} geometry,'' \emph{IEEE Transactions on
  Medical Imaging}, vol.~21, no.~7, pp. 823--828, 2002.

\bibitem{FDCT}
E.~Cand{\`e}s, L.~Demanet, D.~Donoho, and L.~Ying, ``Fast discrete curvelet
  transforms,'' \emph{Multiscale Modeling \& Simulation}, vol.~5, no.~3, pp.
  861--899, 2006.

\bibitem{starck2002curvelet}
J.-L. Starck, E.~J. Cand{\`e}s, and D.~L. Donoho, ``The curvelet transform for
  image denoising,'' \emph{IEEE Transactions on Image Processing}, vol.~11,
  no.~6, pp. 670--684, 2002.

\bibitem{SCIOAD}
D.~L. Donoho, ``Sparse components of images and optimal atomic
  decompositions,'' \emph{Constructive Approximation}, vol.~17, no.~3, pp.
  353--382, 2001.

\bibitem{3DFDCT}
L.~Ying, L.~Demanet, and E.~Candes, ``{3D} discrete curvelet transform,'' in
  \emph{Wavelets XI}, vol. 5914, 2005, p. 591413.

\bibitem{RIPrelaxe}
M.~Fornasier and H.~Rauhut, ``Compressive sensing,'' in \emph{Handbook of
  Mathematical Methods in Imaging}.\hskip 1em plus 0.5em minus 0.4em\relax
  Springer, 2011, pp. 187--228.

\bibitem{enhancing}
E.~J. Cand{\`e}s, M.~B. Wakin, and S.~P. Boyd, ``Enhancing sparsity by
  reweighted $\ell_1$ minimization,'' \emph{Journal of Fourier Analysis and
  Applications}, vol.~14, no. 5-6, pp. 877--905, 2008.

\bibitem{IRLSMSR}
I.~Daubechies, R.~DeVore, M.~Fornasier, and C.~S. G{\"u}nt{\"u}rk,
  ``Iteratively reweighted least squares minimization for sparse recovery,''
  \emph{Communications on Pure and Applied Mathematics}, vol.~63, no.~1, pp.
  1--38, 2010.

\bibitem{SSLS}
S.~Foucart and M.~Lai, ``Sparsest solutions of underdetermined linear systems
  via $\ell_q$-minimization for $0<q\leq1$,'' \emph{Applied and Computational
  Harmonic Analysis}, vol.~26, no.~3, pp. 395--407, 2009.

\bibitem{UMSS}
M.~Lai and J.~Wang, ``An unconstrained $\ell_q$ minimization with $0<q\leq1$
  for sparse solution of underdetermined linear systems,'' \emph{SIAM Journal
  on Optimization}, vol.~21, no.~1, pp. 82--101, 2011.

\bibitem{SALSA}
M.~V. Afonso, J.~M. Bioucas-Dias, and M.~A. Figueiredo, ``Fast image recovery
  using variable splitting and constrained optimization,'' \emph{IEEE
  Transactions on Image Processing}, vol.~19, no.~9, pp. 2345--2356, 2010.

\bibitem{FISTA}
A.~Beck and M.~Teboulle, ``A fast iterative shrinkage-thresholding algorithm
  for linear inverse problems,'' \emph{SIAM journal on imaging sciences},
  vol.~2, no.~1, pp. 183--202, 2009.

\bibitem{ADMM}
S.~Boyd, N.~Parikh, E.~Chu, B.~Peleato, and J.~Eckstein, ``Distributed
  optimization and statistical learning via the alternating direction method of
  multipliers,'' \emph{Foundations and Trends® in Machine Learning}, vol.~3,
  no.~1, pp. 1--122, 2011.

\bibitem{CGLS}
C.~C. Paige and M.~A. Saunders, ``{LSQR}: An algorithm for sparse linear
  equations and sparse least squares,'' \emph{ACM Transactions on Mathematical
  Software}, vol.~8, no.~1, pp. 43--71, 1982.

\bibitem{rullan2018hamilton}
F.~Rullan and M.~M. Betcke, ``Hamilton-{G}reen solver for the forward and
  adjoint problems in photoacoustic tomography,'' \emph{arXiv preprint
  arXiv:1810.13196}, 2018.

\bibitem{rullan2020photoacoustic}
F.~Rullan, ``Photoacoustic tomography: flexible acoustic solvers based on
  geometrical optics,'' Ph.D. dissertation, UCL (University College London),
  2020.

\bibitem{frikel2020sparse}
J.~Frikel and M.~Haltmeier, ``Sparse regularization of inverse problems by
  operator-adapted frame thresholding,'' in \emph{Mathematics of Wave
  Phenomena}.\hskip 1em plus 0.5em minus 0.4em\relax Springer, 2020, pp.
  163--178.

\end{thebibliography}

%








\end{document}